\documentclass[10pt,twocolumn,letterpaper]{article}

\usepackage{wacv}
\usepackage{times}
\usepackage{epsfig}
\usepackage{graphicx}
\usepackage{amsmath}
\usepackage{amssymb}

\usepackage[
backend=biber,
style=ieee,
citestyle=ieee,
]{biblatex}
\AtBeginBibliography{\small}

\usepackage[utf8]{inputenc}
\usepackage[T1]{fontenc}
\DeclareUnicodeCharacter{2212}{-}

\usepackage{xcolor}

\makeatletter
\@namedef{ver@everyshi.sty}{}
\makeatother
\usepackage{tikz}
\usetikzlibrary{shapes}
\usetikzlibrary{automata,topaths}
\usetikzlibrary{calc}
\usepackage{pgfplots}
\pgfplotsset{compat=1.16}

\usepackage{listings}%
\usepackage{minitoc}

\usepackage{csvsimple}

\usepackage{wrapfig}

\usepackage{pdfpages}

\usepackage{afterpage}

\usepackage{rotating}
\usepackage{enumitem}

\usepackage{mathtools}

\usepackage{dsfont}

\usepackage{multirow}
\usepackage{colortbl}

\usepackage[toc,page]{appendix}

\usepackage{url}

\usepackage{caption}

\usepackage{amsbsy}
\usepackage{stmaryrd}

\usepackage{nicefrac}

\usepackage{pifont}%

\usepackage{booktabs}

\def\wacvPaperID{1234} %

\wacvfinalcopy %

\usepackage[breaklinks=true,bookmarks=true,pdfa]{hyperref}

\usepackage{amsmath,amsfonts,bm}

\def\eqref#1{equation~\ref{#1}}

\def\1{\bm{1}}

\DeclareMathAlphabet{\mathsfit}{\encodingdefault}{\sfdefault}{m}{sl}
\SetMathAlphabet{\mathsfit}{bold}{\encodingdefault}{\sfdefault}{bx}{n}

\let\pgfimageWithoutPath\pgfimage 
\renewcommand{\pgfimage}[2][]{\pgfimageWithoutPath[#1]{fig/#2}}

\usepackage{calculator}

\newcommand{\nnetwork}[1]{\mathit{#1}}

\newcommand{\pixtvex}{\nnetwork{R}}
\newcommand{\rti}{\nnetwork{F}_\nnetwork{r2i}}
\newcommand{\itr}{\nnetwork{F}_\nnetwork{i2r}}
\newcommand{\discr}{\nnetwork{D}}
\newcommand{\DR}{\nnetwork{SDR}}
\newcommand{\ncirc}{\,\circ\,}
\newcommand{\layers}[1]{\texttt{#1}}

\newcommand{\imgx}{x}
\newcommand{\iexp}[1]{\mathbb{E}_{\imgx \sim p}\bigl[\; #1 \;\bigr]}

\makeatletter
\renewcommand\paragraph{\@startsection{paragraph}{4}{\z@}%
                                    {1.5ex \@plus1ex \@minus.2ex}%
                                    {-1em}%
                                    {\normalfont\normalsize\bfseries}}
\makeatother

\begin{document}

\title{Style Agnostic 3D Reconstruction via Adversarial Style Transfer}

\author{
Felix Petersen $\quad$ Bastian Goldluecke $\quad$ Oliver Deussen\\
University of Konstanz\\
{\tt\small \{first.last\}@uni-konstanz.de}
\and
Hilde Kuehne\\
University of Frankfurt\\
IBM-MIT Watson AI Lab
}

\maketitle

	\begin{abstract}
		
Reconstructing the 3D geometry of an object from an image is a major challenge in computer vision.
Recently introduced differentiable renderers can be leveraged to learn the 3D geometry of objects from 2D images, but those approaches require additional supervision to enable the renderer to produce an output that can be compared to the input image.
This can be scene information or constraints such as object silhouettes, uniform backgrounds, material, texture, and lighting.
In this paper, we propose an approach that enables a differentiable rendering-based learning of 3D objects from images with backgrounds without the need for silhouette supervision.
Instead of trying to render an image close to the input, we propose an adversarial style-transfer and domain adaptation pipeline that allows to translate the input image domain to the rendered image domain.
This allows us to directly compare between a translated image and the differentiable rendering of a 3D object reconstruction in order to train the 3D object reconstruction network. 
We show that the approach learns 3D geometry from images with backgrounds and provides a better performance than constrained methods for single-view 3D object reconstruction on this task.

	\end{abstract}

	\section{Introduction}
	
	Inferring the 3D geometry of an object from an arbitrary single image is an interesting and challenging task that is easy for humans but still poses a hard computational problem.
	This problem is challenging because, by projecting a 3D shape onto a 2D image, depth information is discarded, and occluded regions are inherently difficult to reconstruct. 
	
	Perhaps the most interesting and promising approaches addressing this problem are those utilizing learning-based methods. %
	These can potentially exploit previously seen examples to overcome missing information such as partial visibility, unknown lighting conditions, etc. 
	Some of these methods rely on 3D shape supervision such as Mem3D~\cite{Yang_2021_CVPR}, DeformNet~\cite{Kurenkov2018}, Pixel2Mesh~\cite{Wang2018}, and 3D-R2N2~\cite{Choy2016}, 
	or key-point correspondences between images~\cite{Kanazawa2018}. %

	Other, less supervised methods predict the geometry using a geometry reconstruction network and projecting it back onto an image space using a differentiable renderer. %
	Such 3D representations can be based on meshes as in case of SoftRas~\cite{Liu2019-SoftRas}, DIB-R~\cite{Chen2019DIB}, and NMR~\cite{Kato2017}, voxels~\cite{Yan2016}, or distance fields as used in SDFDiff~\cite{Jiang2020-SDFDiff}.
	The idea here is to predict a 3D model from a single image, render the silhouette of this model using a differentiable renderer, and impose a loss function that ensures that the rendered silhouette again resembles the silhouette input image.
	This strategy allows learning the respective shape models without the need for annotation of 3D shapes or key-point correspondences between images.
	An interesting problem arising in this context is that renderings need to be compared to input images for the loss function. 
	Thus, the renderer has to produce images that look stylistically similar to the input images to allow for a meaningful comparison.

	This can be challenging for differentiable renderers due to the complexity of images:
	If the renderer does not have enough degrees of freedom, it cannot produce a realistic, matching image.
	However, if it has too many degrees of freedom, trivial solutions such as a background filling the entire image with the texture of the input image become possible, and training becomes unstable.
	Therefore, current approaches require additional supervision and constraints, such as uniform background, known lighting conditions, materials, textures, and  supervision of silhouettes ~\cite{Kato2017, Liu2019-SoftRas, Chen2019DIB, Yan2016, Jiang2020-SDFDiff}.
	
	\usetikzlibrary{positioning}
	\usetikzlibrary{calc}

	\newcommand{\pixtopixpicture}{
		\draw [semithick]  (0, 0.1) -- (2, 0.1);
		\draw [semithick]  (0, 0.2) -- (2, 0.2);
		\draw [semithick]  (0, 0.3) -- (2, 0.3);
		\draw [semithick]  (0, 0.7) -- (2, 0.7);
		\draw [semithick]  (0, 0.8) -- (2, 0.8);
		\draw [semithick]  (0, 0.9) -- (2, 0.9);
		\draw [thick, fill=white]  (0, 0) -- (0, 1) -- (1, 0.65) -- (2, 1) --
		(2, 0) -- (1, 0.35) -- cycle;
	}
	
	\newcommand{\pixtovexpicture}{
		\draw [thick, fill=white]  (0, 0) -- (0, 1) -- (0.8, 0.75) -- (1.4, 0.75) --
		(1.4, 0.25) -- (.8, 0.25) -- cycle;
	}
	
	\newcommand{\sdrpicture}{
		\draw [very thick, fill=white, rounded corners=.5mm] (0.45, 0.4) -- (0, 0.05) --  (0, 0.95) -- (0.45, 0.6);
		\draw [very thick, fill=white, rounded corners=1mm] (1, 0) -- (0.45, 0) |- (1.7, 1) |- (0.45, 0) |- (1, 1);
	}
	
	\newcommand{\selectedImg}{0026100}
	
	\begin{figure*}[t]
		\centering
		\includegraphics[width=0.78\linewidth]{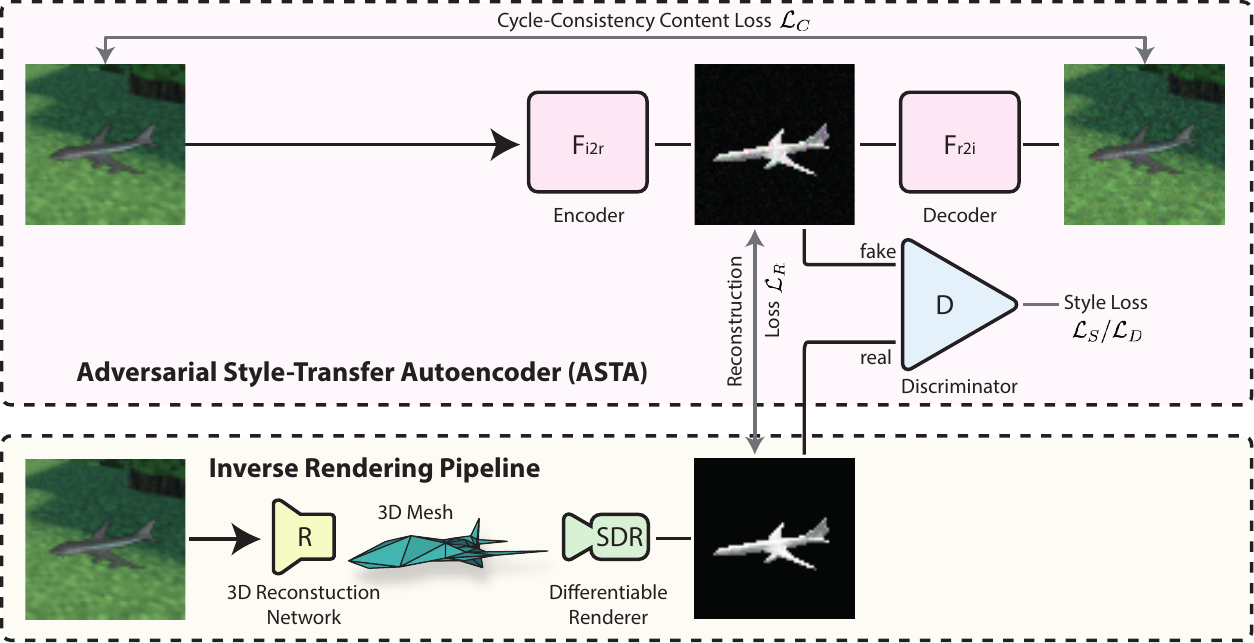}
		\caption{
			Overview of the framework. 
			We extend the differentiable renderer-based 3D geometry reconstruction by an image-to-image domain adaptation network ($\itr$) to allow for training with arbitrary images.
			$\itr$ translates input images into images that could have been renderered.
			We train $\itr$ using an adversarial style loss and a cycle-consistency content loss.
		}
		\label{fig:overview}
	\end{figure*}

	To alleviate those constraints and to deal with the challenge of complex image formation, we propose a framework that combines differentiable rendering based 3D reconstruction with adversarial domain adaptation to bridge the gap between rendered and input images as shown in Fig.~\ref{fig:overview}. 
	Starting from a single input image, we predict the 3D shape of the object with a CNN that converts the image input to a 3D mesh, and apply a differentiable renderer \cite{Liu2019-SoftRas} to produce a 2D rendering of the predicted mesh. 
	To allow for comparison between the rendered representation and the input image, we propose to use an Adversarial Style-Transfer Autoencoder (ASTA) pipeline, which converts the input image to a representation that could have been generated by the renderer. 
	The encoder and decoder are locally restricted U-Net architectures such that the ``latent space'' is a grey-scale image. 
	Thus, the encoder first translates the input image into a representation that resembles the rendered images and the decoder translates this back into its original representation.
    Following the ideas of neural style transfer and adversarial domain adaptation, we use an adversarial disciminator to enforce the ``latent'' space to be stylistically similar to images produced by the differentiable renderer.
    As the autoencoder needs to reconstruct the original image, it preserves the content.
    With this, we can transfer the input image to the differentiable renderer's output space, rather than forcing the renderer to reproduce the input image.

    As only input images and no smoothly rendered images are initially given, we cannot directly train the style-transfer.
	We resolve this by adopting a co-evolutionary strategy to train the networks in an interleaved way, such that the renderings of previous reconstruction guesses are used.

	We first evaluate the proposed framework on the ShapeNet data set~\cite{Chang2015ShapeNet}.
	To demonstrate the proposed framework's abilities, we render objects of ShapeNet with and without random background scenes. 
	We show that, for both cases, the proposed framework learns the 3D object shape.
	For images with backgrounds, we outperform other methods even if they are provided additional silhouette supervision, e.g., by weakly-supervised segmentation.
	On the ShapeNet data set with textured backgrounds, we achieve a mean intersection-over-union (IoU) of $0.3816$, whereas the baseline with additional silhouette supervision achieves a mean IoU of $0.2563$.
	Second, we apply the model trained on ShapeNet to photos of LEGO models of ShapeNet object classes, demonstrating that the method can reproduce the performance on the synthetic images also in this case.

	\paragraph{Contributions.}
	We present a framework for unsupervised training of 3D reconstruction from a single image with random background scenes.
	We propose using an  adversarial style-transfer autoencoder as image-to-image domain translator, which converts input images into images that could have been rendered by a smooth differentiable renderer.
	We train this domain translator to preserve the content while changing the style and---by that---bridge the gap between arbitrary images and the constraints of differentiable renderers.
    As the rendering domain is initially unknown, we leverage a co-evolutionary training of domain adaptation.
	In particular, we can relax assumptions on and do not require supervision of materials, colors, lighting, background, and silhouette---they do not need to be controlled nor constant across the data set.

	\section{Related Work}
	Since our goal is unsupervised learning of single-view 3D reconstruction, we first focus on related work based on differentiable rendering.
	In addition, we present related work on style-transfer, adversarial domain adaptation, and image-to-image translation.
	\paragraph{3D Object Reconstruction via Differentiable Renderers.}
	Over the past years, many differentiable renderers for 3D meshes have been presented \cite{Kato2017, Loper2014, Li2018, Henderson2018, Delaunoy2011, Ramamoorthi2001, Meka2018, Athalye2017, Richardson2017, Liu2019-SoftRas, Loubet2019ReparameterizingRendering, Kundu2018}.
	Those renderers are differentiable with respect to lighting, geometry \cite{Li2018, Loper2014, Kato2017, Delaunoy2011, Richardson2017, Liu2019-SoftRas, Loubet2019ReparameterizingRendering, Chen2019DIB, Henderson2018, Jiang2020-SDFDiff},
	material \cite{Ramamoorthi2001, Meka2018, Athalye2017, Loubet2019ReparameterizingRendering, Chen2019DIB}, or texture \cite{Ramamoorthi2001, Athalye2017, Loubet2019ReparameterizingRendering, Chen2019DIB}.
	Using a differentiable renderer, a mesh optimization process can be performed to reconstruct a single
	3D object~\cite{Delaunoy2011, Li2018, Loper2014, Kato2017, Loubet2019ReparameterizingRendering, Chen2019DIB}.
	For that, an initial mesh can be optimized iteratively by adapting its geometry guided by back-propagated gradients.
	With state-of-the-art deep learning techniques and 3D supervision, high-quality 3D object reconstruction can be done via direct prediction~\cite{Achlioptas2017, Choy2016, Fan2016, Jiang2018, Kurenkov2018, Wang2018}.
	This approach is orders of magnitude faster than optimization and allows transferring knowledge to unseen examples and occluded areas.
	For 3D unsupervised 3D object prediction, state-of-the-art are encoder-renderer architectures where the encoder is trained for 3D mesh prediction
	and the differentiable renderer plays the role of a decoder.
	Thus, the encoder-renderer network predicts the 3D model in the latent space~\cite{Henderson2018, Richardson2017, Liu2019-SoftRas, Kato2017, Chen2019DIB}.
	Zhang~\textit{et al.}~\cite{zhang2021-image-gan-diff-render} extend this idea by integrating GANs to generate alternative views for each input image, while also relying on silhouette supervision.
	These encoder-render methods are restricted to reconstructing from images that look like images produced by the integrated
	differentiable renderer or require additional silhouette supervision, which limits their application.

	The first to raise the idea of learning from 2D images without silhouette supervision were Henderson~\etal~\cite{Henderson_2020_CVPR}. 
	They performed textured 3D mesh generation from images without silhouette annotation via a 2D VAE based framework.
	The difference is that they focus on mesh generation, while our focus lies on mesh reconstruction.
	To this end, they focus on making the rendering equal to the input image, while we focus on translating the input image into a rendered image.

	\paragraph{Style-Transfer and Adversarial Domain Adaptation.}
	Neural Style-Transfer (NST) has been first proposed by Gatys~\etal~\cite{Gatys2015-NST}, where the activations of a pretrained CNN are used for the style loss, allowing to optimize the style-transfer on a single image.
	Using a network to predict the style-transferred image, Isola~\etal~\cite{Isola2016-Pix2Pix} propose the image-to-image translation architecture Pix2Pix, training it through a Generative Adversarial Network (GAN).
	This allows for fast inference as it shifts the process from optimization to prediction.
	Zhu~\etal~\cite{Zhu2017-CycleGAN} extend this idea by introducing two GANs, translating an image as well as back-translating it, and thus including a cycle-consistency loss (CycleGAN).
	Based on this idea, adversarial approaches to unsupervised domain adaptation have been presented \cite{hoffman2018cycada, wilson2020survey, tzeng2017adversarial}.
	Hoffman~\etal~\cite{hoffman2018cycada} propose extending CycleGAN by a semantic loss and a feature loss for adversarial domain adaptation.
	We propose a variation of these ideas, where an image-to-image domain adaptation network translates an input image to an image from the domain of the encoder-rendering pipeline's outputs and preserves the content by a cycle-consistency through autoencoding.

	\begin{figure*}[t]
		\centering
		\includegraphics[width=.48\linewidth]{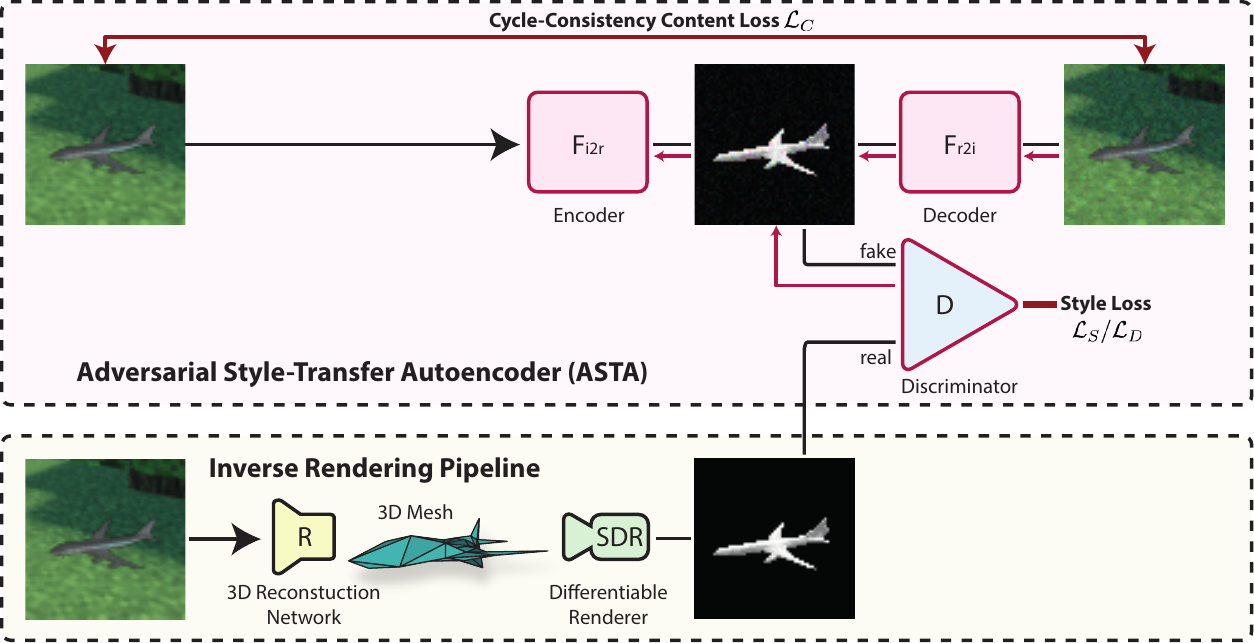}\hfill
		\includegraphics[width=.48\linewidth]{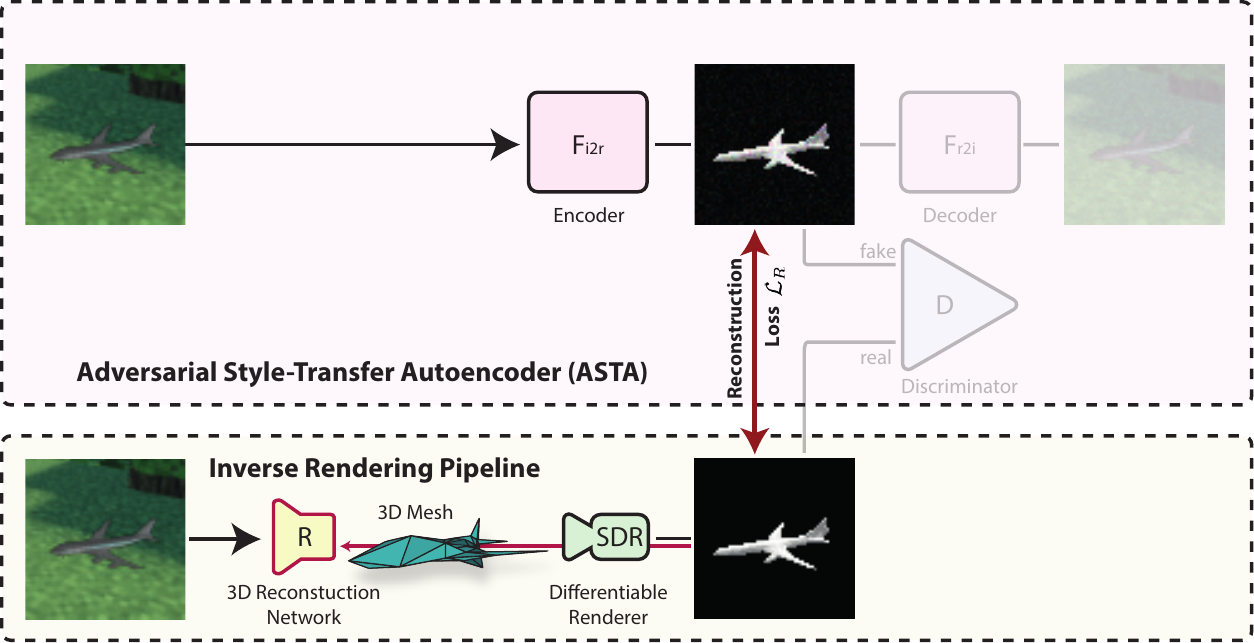}
		\caption{
			Training procedure of the proposed framework.
			Left: in the first stage, we train the domain adaptation with the cycle-consistency content loss as well as the adversarial style loss.
			Right: in the second stage, we train the reconstruction network with the geometry reconstruction loss. 
			The red arrows indicate back-propagation of gradients.
		}
		\label{fig:overview-training}
	\end{figure*}

	\section{\hbox{Style-Agnostic Unsupervised Reconstruction}}
	To perform an unsupervised style-agnostic reconstruction, we need to bridge the gap between input images
	and the output of the renderer.
	Our architecture is divided into two main parts as shown in Fig.~\ref{fig:overview}.
	The image-to-image translation pipeline (top) translates input images into images which resemble those generated by the differentiable renderer.
	The reconstruction-rendering pipeline (bottom) consists of the 3D reconstruction network $\pixtvex$ and a smooth differentiable renderer $\DR$, forming together the encoder-renderer network.
	The reconstruction network translates an image input into a 3D geometry representation, which is then fed into the differentiable renderer to produce the respective rendered representation of the 3D geometry.
	Both parts are trained in an alternating fashion.
	In the remainder of this section, we discuss both pipelines in detail, introduce the necessary loss functions,
	and describe the overall training strategy to tie the components together.
	
	\subsection{Loss Functions for the Domain Adaptation}
	\label{sub:objective}
	
	The network for image-to-image translation aims to translate images~$\imgx$ drawn from the input image domain~$p$ into images that look like they have been rendered by the differentiable renderer~$\DR$.
	
	Training of this image-to-rendering network~$\itr$, is controlled by two loss functions, an adversarial style loss arising from a discriminator between renderings and pseudo renderings, and a cycle-consistency loss based on the inverse mapping of the pseudo-rendered image back to the original image. 
	Each sub-network is described in detail in the following. 
	
	\paragraph{Adversarial Style Loss.}
	To convert input images into their rendered representations, we need to transfer the style from rendered images to the input images. 
	To this end, we use the adversarial loss of a discriminator~$\discr$, which discriminates between images that have actually been rendered and images created by~$\itr(\imgx)$, where~$\imgx$ is an input image.
	Different from typical adversarial losses as, e.g., used in pix2pix~\cite{Isola2016-Pix2Pix} or CycleGAN~\cite{Zhu2017-CycleGAN}, we can not sample from the domain of smoothly rendered images. 
	Instead, we use those images generated by the reconstruction network and the differentiable renderer as examples of rendered objects, as we do not know the geometry of the 3D objects in advance.
	Note that this requires a co-evolutionary training of the reconstruction network~$\pixtvex$ and the adversarial style transfer, as detailed in \ref{sec:training-strategy}. 

	Our loss function $\mathcal{L}_D$ for the discriminator~$\discr$ and style loss $\mathcal{L}_S$ for the image-to-rendering encoder~$\itr$ are:
	\begin{align}
		\mathcal{L}_{\discr} =&\ \iexp{
			\log\discr (\itr(\imgx)) \;+\; 
			\notag \\ &\qquad\quad
			\log (1 - \discr( \DR ( \pixtvex(\imgx) ))) } 
		\label{eq:loss-d}
		\\
		\mathcal{L}_S =&\ \iexp{ \log (1 - \discr( \itr( \imgx ))) }
		\label{eq:loss-s}
	\end{align}
	
	To ensure that, for each domain adaptation, the predicted rendering does not differ too much from the current rendered prediction, we include the loss term
	\begin{equation}
		\mathcal{L}_J = \iexp{ \max( 0, \delta - J(\DR ( \pixtvex(\imgx)),\, \itr(\imgx) )) }
		\label{eq:loss-j}
	\end{equation}
	to encourage that the Jaccard similarity coefficient~$J$ is larger than $\delta=0.25$.
	We relax the Jaccard similarity to $J(x, y) = \frac{x y}{x + y - xy}$ as per probabilistic real-valued logic.

	\paragraph{Cycle-Consistency Content Loss.} 
	We train the autoencoder with a cycle-consistency content loss to preserve the content during the image-to-rendering transfer.
	For that, we use the rendering-to-image decoder termed~$\rti$, which is trained to compute the inverse map of $\itr$.
	We find that, as the autoencoder needs to preserve information, the easiest path (and thus the path chosen) is to keep the contents because it needs to reconstruct the input again from the stylized image. 
	As experiments by Chu~\etal~\cite{Chu2017CycleGANMasterSteganography} indicate, a cycle-consistency can be fulfilled even while the inner image $\itr(\imgx)$ is almost constant because most information can be hidden in high-frequency detail.
	To reinforce that the content is preserved in the ``latent'' / smoothly rendered space, we apply random noise to the latent space before feeding it into the decoder to prevent it from storing all information in high-frequency details.
	Thus, we add the noise term $\epsilon$ to the input of the rendering-to-image decoder~$\itr(\imgx)$.
	We define the cycle-consistency loss, which is applied to both, the image-to-rendering and the rendering-to-image translation networks $\itr$ and $\rti$ , as
	\begin{equation}
		\mathcal{L}_C = \iexp{ \| \imgx - \rti( \itr(\imgx) + \epsilon )\|_h },
		\label{eq:loss-c}
	\end{equation}
	where $\|\cdot\|_h$ denotes the (smooth) Huber norm
	and $\epsilon \sim \mathcal{N}(0, \sigma_{\mathrm{noise}}^2)$ where $\sigma_{\mathrm{noise}}=0.15$.

	In total, the style, content, and Jaccard losses for $\itr$ and $\rti$ accumulate to
	\begin{equation}
		\begin{aligned}
			\mathcal{L}_{\itr} =& \mathcal{L}_C + \mathcal{L}_S + \mathcal{L}_J\\
			\mathcal{L}_{\rti} =& \mathcal{L}_C 
			.
		\end{aligned}
	\end{equation}
	
	\subsection{Loss Function for the Geometry Reconstruction}
	\label{sec:loss-reconstruction}
	To train the reconstruction network, we consider the difference between the rendered predictions for the reconstruction and the stylized input images, as shown in Fig.~\ref{fig:overview-training}. 
	Following Liu~\etal~\cite{Liu2019-SoftRas} and Kato~\etal~\cite{Kato2017}, we sample two input images of the same object, using one of them as input to the reconstruction network, whereas the second view is only used to validate the result from an alternative view point.
	The reconstruction loss is therefore defined as: 
\begin{align}
		\label{eq:loss-r}
		\mathcal{L}_{\pixtvex} = 
		\mathbb{E}_{\imgx, y \sim p}\bigl[\; &\DR_x ( \pixtvex( \imgx ) ) \,-\, \itr( x )\|_h + \notag\\
		&\DR_y ( \pixtvex( \imgx ) ) \,-\, \itr( y )\|_h\;\bigr]
\end{align}
	Here $x,y\sim p$ are pairs of images of the same object, $\DR_x$ renders the object from the view point of $x$, and $\DR_y$ renders the object from the alternative view point of $y$.
	To guide the prediction towards non-degenerate and reasonably shaped meshes,
	we also apply regularization losses as done by Liu~\etal~\cite{Liu2019-SoftRas}.

	\begin{figure*}[t]
		\centering
		\includegraphics[width=\linewidth]{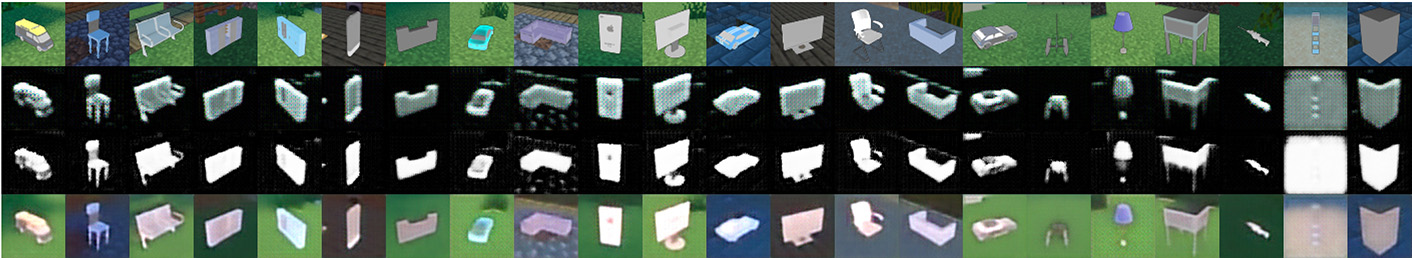}
		\caption{
			Examples of the domain adaptation in the tenth training cycle.
			The top row displays the input image $\imgx$, the second the RGB channels of $\itr(\imgx)$, the third the alpha channel of $\itr(\imgx)$, and the bottom row displays $\rti(\itr(\imgx))$.
		}
		\label{fig:style-transfer-examples}
	\end{figure*}

	\subsection{Overall Training Strategy}
	\label{sec:training-strategy}
	The challenge in training is that, to train reconstruction, we need to compare the smoothly rendered images of the predicted shape to the input, which requires domain adaptation.
	But, to train the domain adaptation, we need to know what a smoothly rendered image looks like.
	We resolve this problem by adopting a co-evolutionary strategy and by training the networks in an interleaved way.
	For each training cycle, we start with training the domain adaptation networks using the respective
	losses~(see Eq.~\ref{eq:loss-s}, Eq.~\ref{eq:loss-j} and Eq.~\ref{eq:loss-c})
	for a fixed number of iterations, and after that train the reconstruction network~$\pixtvex$ using the reconstruction loss (see Eq.~\ref{eq:loss-r}).
	Note that we start with a setting where the reconstruction is a constant sphere, a basic domain adaptation is learned, which then initializes the reconstruction network.
	For each new training cycle, we keep the weights of the reconstruction network, but re-initialize the domain adaptation and continue the training as described above.
	This re-initialization allows the domain adaptation to be trained with the current optimal guesses for the domain of rendered objects, and to recover from previous errors induced by erroneous reconstructions.
	In the next training cycle, training of the domain adaptation improves since the rendered reconstructions are more representative samples of the rendered image domain.
	Examples for the domain adaptation in the tenth training cycle for various objects are shown in Fig.~\ref{fig:style-transfer-examples}.

	\setlength{\tabcolsep}{4pt}
	\begin{table*}[t]
		\caption{
			Mean 3D IoU for 13 object classes from ShapeNet \cite{Chang2015ShapeNet} averaged over 3 runs. 
			We report baselines for silhouette supervision, uniform backgrounds, and textured background. 
			In both cases (of the latter), domain adaptation significantly improves performance, and for uniform backgrounds our method outperforms older methods that require silhouettes.
			\label{tab:shapenet}
		}
		\resizebox{\textwidth}{!}{
			\centering
			\begin{tabular}{lcccccccccccccc} 
				\toprule
				Method    & Airplane & Bench  & Dresser & Car    & Chair  & Display & Lamp & Speaker & Rifle & Sofa & Table & Phone & Vessel & \textit{Mean}       \\ 
				\bottomrule
				\toprule
				\textbf{With} silhouette supervision \\
				\midrule
				Yan~\etal~\cite{Yan2016} (retrieval)                      & 0.5564   & 0.4875 & 0.5713  & 0.6519 & 0.3512 & 0.3958  & 0.2905 & 0.4600   & 0.5133 & 0.5314  & 0.3097 & 0.6696 & 0.4078  & 0.4766       \\
				Yan~\etal~\cite{Yan2016} (voxel)                          & 0.5556   & 0.4924 & 0.6823  & 0.7123 & 0.4494 & 0.5395  & 0.4223 & 0.5868   & 0.5987 & 0.6221  & 0.4938 & 0.7504 & 0.5507  & 0.5736       \\
				Chen~\etal~\cite{Chen2019DIB} (DIB-R)                     & 0.570    & 0.498  & \bf{0.763}   & \bf{0.788}  & 0.527  & 0.588   & 0.403  & \bf{0.726}    & 0.561  & 0.677   & 0.508  & 0.743  & 0.609   & 0.612        \\
				Kato~\etal~\cite{Kato2017} (NMR)                          & 0.6172   & 0.4998 & 0.7143  & 0.7095 & 0.4990 & 0.5831  & 0.4126 & 0.6536   & 0.6322 & 0.6735  & 0.4829 & 0.7777 & 0.5645  & 0.6015       \\
				Liu~\etal~\cite{Liu2019-SoftRas} (SoftRas (sil.))          & 0.6419   & 0.5080 & 0.7116  & 0.7697 & 0.5270 & 0.6156  & \bf{0.4628} & 0.6654   & \bf{0.6811} & 0.6878  & 0.4487 & 0.7895 & 0.5953  &  0.6234       \\ 
				Liu~\etal~\cite{Liu2019-SoftRas} (SoftRas (sil.+color))                & \bf{0.6670}   & \bf{0.5429} & 0.7382  & 0.7876 & \bf{0.5470} & \bf{0.6298}  & 0.4580 & 0.6807   & 0.6702 & \bf{0.7220}  & \bf{0.5325} & \bf{0.8127} & \bf{0.6145} & \bf{0.6464}     \\
				\bottomrule
				\toprule
				\textbf{Without} silhouette supervision \\
				\midrule
				\textbf{Uniform backgrounds} \\
				SoftRas  \textbf{(no sil. supervision)}                             & 0.1429 & 	0.1112 & 	0.2342 & 	0.3341 & 	0.1589 & 	0.1713 & 0.0837 & 0.1916 & 	0.0769 & 0.2125 & 	0.0830 & 	0.1918 & 	0.1644 & 	0.1659 \\
				\textit{(Ours)} Domain Adaptation \textbf{(no sil. supervision)}    & \textbf{0.5502} & 	\textbf{0.4441} & 	\textbf{0.6538} & 	\textbf{0.7097} & 	\textbf{0.4995} & 	\textbf{0.5756} & 	\textbf{0.4250} & \textbf{0.6337} & 	\textbf{0.6082} & 	\textbf{0.6543} & 	\textbf{0.4475} & 	\textbf{0.7178} & 	\textbf{0.5431} & 	\textbf{0.5740} \\
				\midrule
				\textbf{Minecraft backgrounds} \\
				SoftRas \textbf{(no sil. supervision)}            & 0.0251 & 0.0409 & 0.1618 & 0.0885 & 0.1111 & 0.0999 & 0.0437 & 0.2481 & 0.0124 & 0.1233 & 0.0836 & 0.0542 & 0.0337 & 0.0866  \\
				SoftRas \textbf{(sil. generated by SEAM)}         & 0.2905 & 0.2032 & 0.0843 & 0.5195 & 0.1626 & 0.2732 & 0.1788 & 0.1558 & \textbf{0.2573} & 0.2257 & 0.2299 & 0.3778 & \textbf{0.3732} & 0.2563     \\
				\textit{(Ours)} Domain Adaptation \textbf{(no sil. supervision)}          & \textbf{0.2994} & \textbf{0.2563} & \textbf{0.5002} & \textbf{0.6078} & \textbf{0.3222} & \textbf{0.4135} & \textbf{0.2480} & \textbf{0.4208} & 0.2551 & \textbf{0.4861} & \textbf{0.2455} & \textbf{0.5485} & 0.3569 & \textbf{0.3816}  
				\\
				\bottomrule
		\end{tabular}}
	\end{table*}
	\setlength{\tabcolsep}{1.4pt}
	
	\section{Experiments\protect\footnote{Our implementation as well as our new data sets are openly available at \href{https://github.com/Felix-Petersen/style-agnostic-3d-reconstruction}{\color{blue!50!black}{github.com/Felix-Petersen/style-agnostic-3d-reconstruction}}.}}

	\begin{figure}
		\centering
		\includegraphics[width=.925\linewidth, trim={0 0cm 0 0cm},clip]{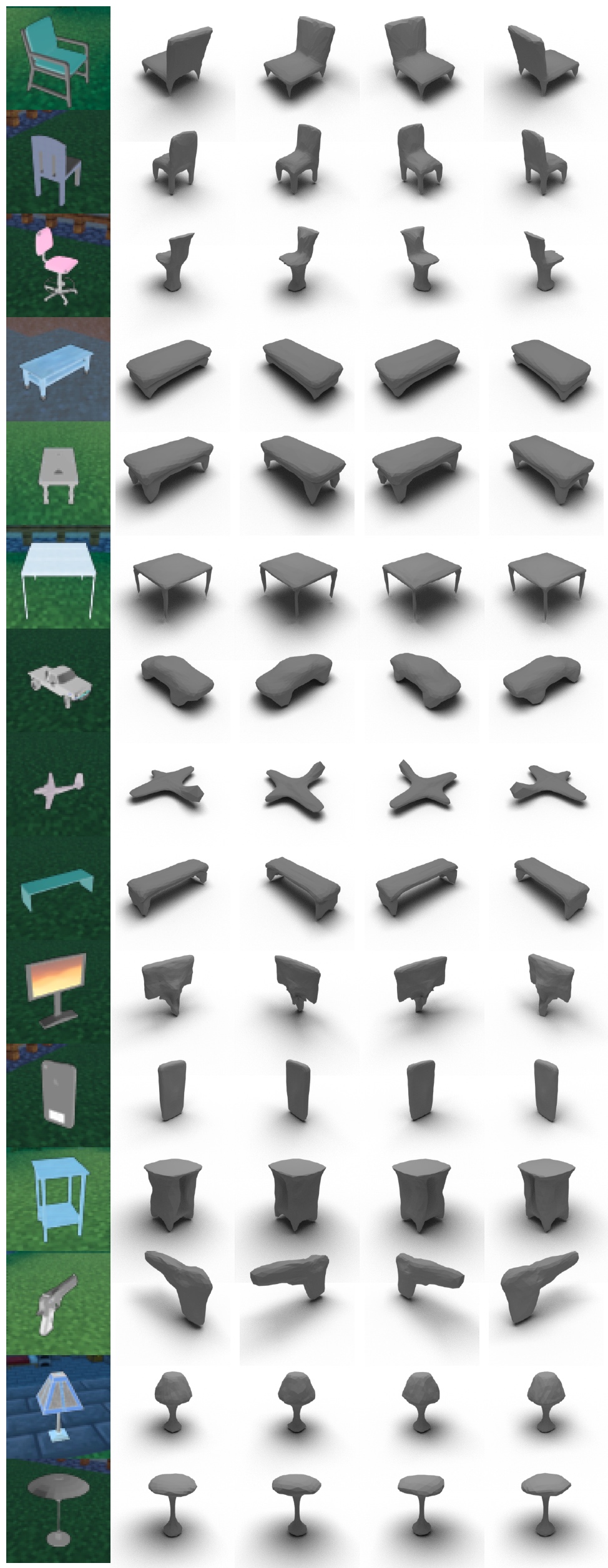}\\[-.5em]
		\caption{
			Examples of the Geometry Reconstruction from single images with Minecraft backgrounds.
			For further examples, see the supplementary material.
		}\vspace{-1.5em}
		\label{fig:reconstruction-examples}
	\end{figure}

	\subsection{Data Sets and Evaluation Metric}
	\label{sec:dataset_metric}
	\paragraph{ShapeNet.} We evaluate the proposed approach on the ShapeNet~\cite{Chang2015ShapeNet} data set.
	ShapeNet is a 3D object data set that allows us to render sample images with uniform as well as textured background and to quantitatively compare the resulting 3D reconstruction to the 3D ground truth shape. For the comparison to state-of-the-art as well as the ablation study, we use the 13 standard benchmark classes from ShapeNet and which have been rendered by Kato~\etal~\cite{Kato2017} at a resolution of $64 \times 64$ from $24$ azimuths at an altitude of $30^{\circ}$ using Blender and are publicly available.

	To show that our method is style-agnostic, we add uniform as well as textured backgrounds to the input and training images of the data set.
	To this end, we render backgrounds of 135 Minecraft scenes with each 3 lighting settings and from 24 azimuths with Blender and combine them with the rendered object views.
	Samples of these images with backgrounds can be found in Fig.~\ref{fig:style-transfer-examples},~\ref{fig:reconstruction-examples},~and~\ref{fig:seam-segmentation-examples}. 
	Note that Fig.~\ref{fig:style-transfer-examples} and~\ref{fig:seam-segmentation-examples} are from the training sets and Fig.~\ref{fig:reconstruction-examples} is from the test set.
	
	We follow the evaluation metric proposed by Kato~\etal~\cite{Kato2017} and used by Liu~\etal~\cite{Liu2019-SoftRas} and Chen~\etal~\cite{Chen2019DIB} and use respective occupancy voxel grids of resolution $32 \times 32 \times 32$ and 3D intersection-over-union (IoU). 
	We report all results based on mean 3D IoU averaged over 3 runs for 13 object classes as well as the mean over all classes.
	\paragraph{ShapeNet2Lego.} To evaluate the model on camera-captured photos, we built seven physical LEGO models and their CAD 3D model from ShapeNet object classes. 
	We captured 12 photos from different angles as shown in Fig.~\ref{fig:lego-recon} with backgrounds consisting of LEGO baseplates and walls. 
	We use this as a second test set and apply the models trained on the synthetic ShapeNet images on those photos as well. 
	
	\subsection{Implementation details}
	\label{sec:implementation}
	To allow comparability with previous works, we build on the source code by Liu~\etal~\cite{Liu2019-SoftRas}.
	This setting also follows the setup by Kato~\etal~\cite{Kato2017}.
	We use the same reconstruction network topology and train with a batch size of $64$, a learning rate of $10^{-4}$, and a uniform sphere with $642$ vertices as base model.
	We further use---wherever applicable---the same hyperparameters, metrics, data set, and seeds to produce comparable results.
	During training of the reconstruction network, we follow the same strategy, i.e., we sample two input images of the same object scene, using one of them as input to the reconstruction network, whereas the second view is only used to validate the result. 
	This is possible as the camera position is given, such that the predicted object geometry can be rendered from both perspectives.
	Note that we demonstrate in Section~\ref{sec:variance-of-azimuth} that, even under strong perturbation of the given camera position by Gaussian noise with a standard deviation of up to $5^{\circ}$, the reconstruction quality does not drop significantly.

	The domain adaptation is re-trained at the beginning of each training cycle.
	We weight the adversarial game in the style loss with a factor of 1 and the cycle-consistency loss of the transfer with a factor of 400. %
	All reported accuracies are averaged over three runs.

	\subsection{Comparison to State-of-the-Art}
	\label{sec:quantitative-results}
	
	\begin{figure}
		\centering
		\includegraphics[width=\linewidth]{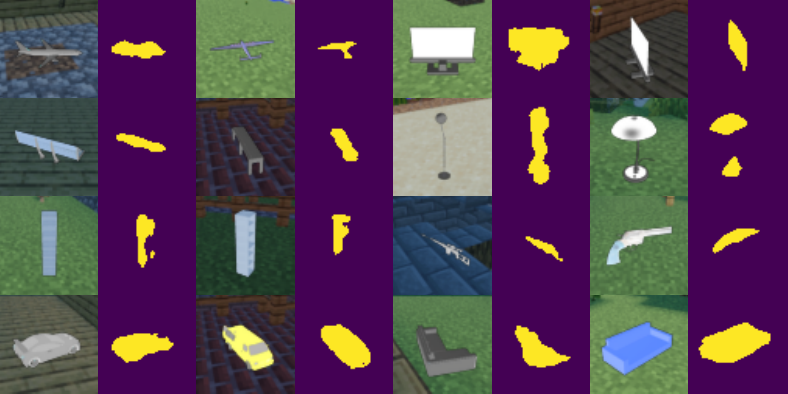}
		\caption{Examples of the weakly-supervised semantic segmentation with SEAM~\cite{Wang_2020_CVPR_SEAM}}
		\label{fig:seam-segmentation-examples}
	\end{figure}
	First, we consider how our method improves existing differentiable rendering pipelines under weaker supervision.
	To this end, we present results of current methods with silhouette supervision and use the best performing differentiable renderer, i.e., SoftRas \cite{Liu2019-SoftRas}, for our method.
	To show the impact of our method, we consider two scenarios. 
	In the first case we render all objects with uniform backgrounds and compare the performance with and without domain adaptation.
	For a second experiment we use images with textured background as described in Sec.~\ref{sec:dataset_metric} and shown in Fig.~\ref{fig:style-transfer-examples} and~\ref{fig:reconstruction-examples}.  
	The results of all settings are shown in Tab.~\ref{tab:shapenet}.
	
	As we use the same differentiable rendering pipeline as Liu~\etal~(SoftRas)~\cite{Liu2019-SoftRas} but with weaker supervision, we consider the performance with silhouette supervision as an upper bound for learning without silhouette supervision. 

	We first compare the performance in the case of uniform backgrounds.
	Comparing baselines with and without silhouette supervision shows that accuracy of the baselines drops significantly when using uniform backgrounds and thus no silhouette supervision, as it is not designed to operate in this setting in general. 
    With our domain adaptation, we reach an accuracy of $57.4\%$ compared to the $64.6\%$ upper bound under stronger supervision and even exceeds the voxel-based perspective transformer networks~\cite{Yan2016} as well as the retrieval baseline method~\cite{Yan2016}. 
	
	Second, we compare the performance in the case of textured Minecraft backgrounds to two baselines based on the SoftRas~\cite{Liu2019-SoftRas} algorithm:
	For the first baseline, we apply the algorithm as is without silhouette supervision, and for the second baseline, we provide a silhouette based on semantic segmentation. 
	To this end, we leverage a state-of-the-art weakly-supervised semantic segmentation method (SEAM)~\cite{Wang_2020_CVPR_SEAM} to generate silhouettes from the rendered images with textured backgrounds. 
	We train the segmentation model on all samples of each object class with the respective class label and apply the resulting model to all samples to get the final silhouettes. 
	Examples of the resulting segmentation are displayed in Fig.~\ref{fig:seam-segmentation-examples}.
	The respective segmentation map is then used as silhouette supervision for the SoftRas algorithm.
	SoftRas (sil. generated by SEAM) is, apart from not using ground truth silhouettes, equivalent to SoftRas (sil.)~\cite{Liu2019-SoftRas}. 
	
	Without silhouette information, the baseline achieves a mean accuracy of $0.0866$ on images with textured backgrounds, which is a significant drop compared to silhouette supervision.  %
	Compared to that, the baseline based on silhouettes generated by the weakly-supervised semantic segmentation method achieves a mean IoU of $0.2563$.
	With the proposed domain adaptation for style-agnostic reconstruction, we achieve a mean IoU of $0.3816$. 
	Overall, the domain adaptation outperforms supervision by weakly-supervised silhouette predictions in eleven out of thirteen categories.
	We show qualitative results in Fig.~\ref{fig:reconstruction-examples}.

	\subsection{Ablation Study}
	We evaluate the impact of the different elements of the network for images with textured background in an ablation study shown in Tab.~\ref{table:ablation-study}.
	As expected, without the cycle-consistency content loss or  the adversarial style loss, the accuracy drops significantly.
	Note that the Jaccard loss has the special role of preventing the domain-adapted image from differing by a too large factor from the rendered reconstruction.
	We found that, without the Jaccard loss, the entire architecture becomes less stable; e.g., one out of three test runs without the Jaccard loss achieves a mean IoU of only $0.0166$.
	Without the domain adaptation, i.e., assuming that the rendered images should be the same as the input images, the mean IoU is $0.0866$.
	
	\setlength{\tabcolsep}{4pt}
	\begin{table}
		\centering 
		\caption{
			Ablation Study.
			We evaluate the impact of all proposed losses as well as the entire domain adaptation.
		}
		\vspace{-.5em}
		\small
		\begin{tabular}{lc} 
			\toprule
			Method & IoU~Acc.  \\ 
			\midrule
			Style-Agnostic Reconstruction & 0.3816  \\ 
			\midrule
			Without Content-Loss (Eq.~\ref{eq:loss-c}) & 0.2293  \\ 
			Without Style-Loss (Eq.~\ref{eq:loss-s}) & 0.2439  \\ 
			Without Jaccard-Loss (Eq.~\ref{eq:loss-j}) & 0.1852  \\
			\midrule
			Without Domain Adaptation & 0.0866  \\ 
			\bottomrule
		\end{tabular}
		\label{table:ablation-study}
	\end{table}
	\setlength{\tabcolsep}{1.4pt}

	\setlength{\tabcolsep}{4pt}
	\begin{table}[h]
		\centering 
		\caption{
			Standard deviation of angles added to the azimuth. 
		}
		\vspace{-.5em}
		\small
		\begin{tabular}{lcccccccccc} 
			\toprule
			Std. & $0^{\ncirc}$ & $1^{\ncirc}$ & $5^{\ncirc}$ & $15^{\ncirc}$ & $25^{\ncirc}$ \\ 
			\midrule
			IoU~Acc.  & 0.3816 & 0.3688 & 0.3726 & 0.2958 & 0.2786  \\
			\bottomrule
		\end{tabular}
		\label{table:variance-of-azimuth}
	\end{table}
	\setlength{\tabcolsep}{1.4pt}

	\subsection{Variation of the Azimuth}
	\label{sec:variance-of-azimuth}
	
	As the azimuth, in ours as well as previous methods, is supervised, we perform an experiment challenging the requirement of azimuth supervision. 
	The motivation of this experiment is to examine whether the proposed method would also be able to work in settings where the camera extrinsics are computed from point correspondences and could be noisy.
	To simulate this setting, we add values drawn from a normal distribution to the ground truth azimuth values. 
	We consider standard deviations of up to $25^{\ncirc}$.
	The results are shown in Tab.~\ref{table:variance-of-azimuth}.
	We find that, up to a standard deviation of $5^\circ$, there is only a small performance penalty indicating that an approximate azimuth is sufficient. %
	
	\newcommand{\iouformat}[1]{$~~0.#1\,~~$}
	\begin{figure*}[t]
	    \centering
	    \includegraphics[width=\linewidth]{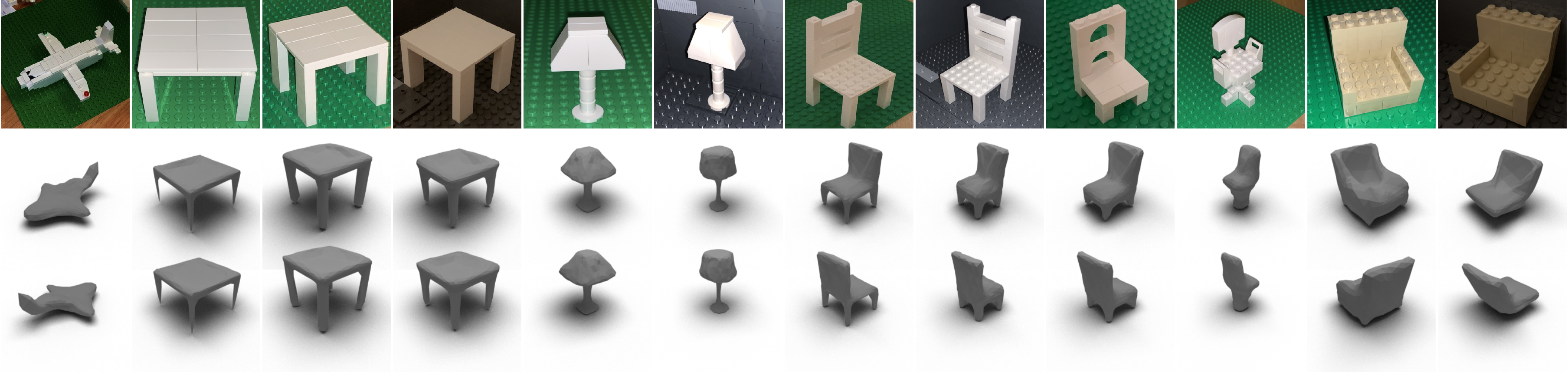}
	    \vspace{-.5em}\\\kern-.5em
	    \iouformat{3246} \iouformat{3397} \iouformat{5543} \iouformat{2506}
	    \iouformat{3798} \iouformat{3333} \iouformat{3255} \iouformat{1859}
	    \iouformat{2541} \iouformat{3346} \iouformat{4244} \iouformat{2725}
	    \kern-.5em\\[-.2em]
	    \caption{3D reconstructions from real photos of LEGO objects. The metric is 3D IoU and the average IoU is $0.3316$.}\vspace{-.3em}
	    \label{fig:lego-recon}  
	\end{figure*}

	\subsection{Behavior of the Domain Adaptation}
	
\begin{figure}[t]
		\centering
		\includegraphics[width=\linewidth]{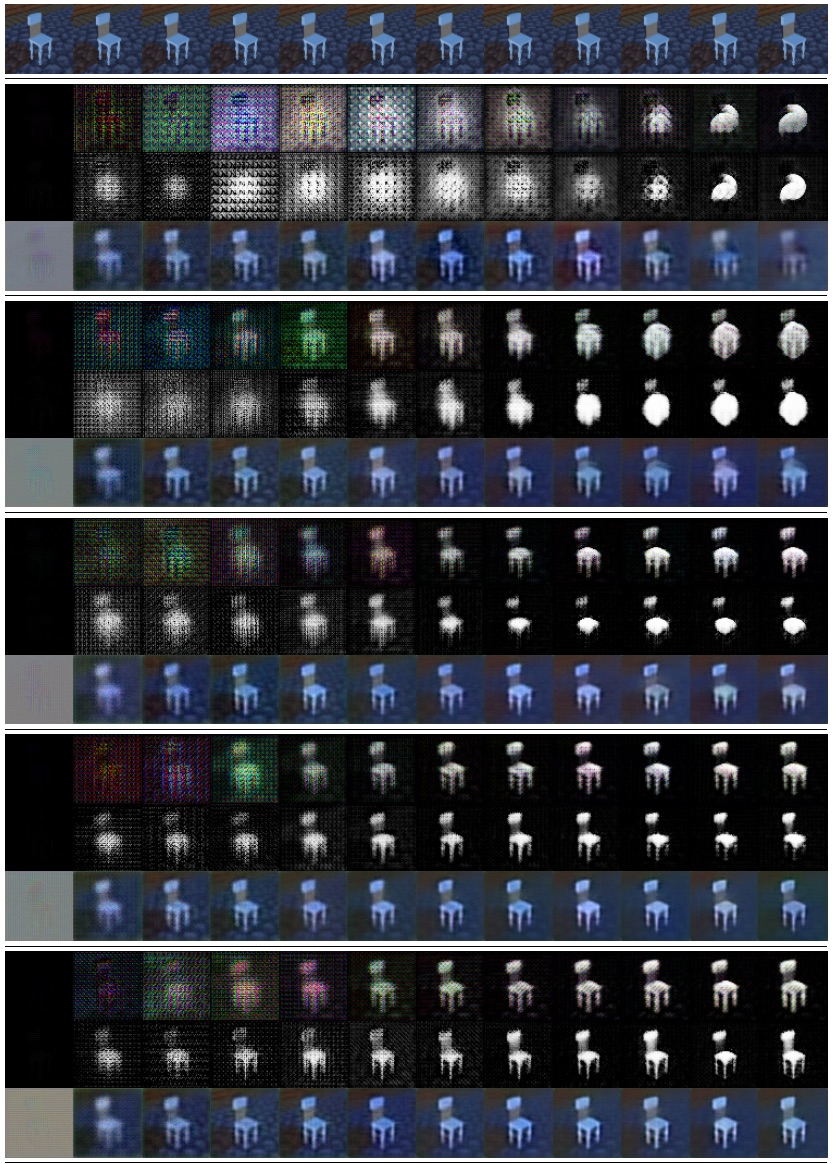}
		\caption{
			Training of the adversarial domain adaptation. 
			The top row is the input (repeated for visual unity.)
			For each block, the top row is $\itr(\imgx)$, the middle row is the silhouette of $\itr(\imgx)$, and the bottom row is the image translated back to the original $\rti(\itr(\imgx))$.
			Each of the five blocks represents one of the first five training cycles (from top to bottom).
			On the horizontal axis, training over $2\,500$ steps is displayed (from left to right).
			For further examples, see Supplementary Material \ref{sm:style-transfer-examples}.
		}
		\label{fig:style-transfer-training-progress}
		\vspace{-2.5em}
	\end{figure}

	In Fig.~\ref{fig:style-transfer-training-progress}, we display the training process for the first 5 training cycles.
	In each training cycle, after training the domain adaptation, the geometry reconstruction network learns to work in harmony with the domain adaptation.
	Over the course of many training cycles, the domain adaptation gets closer to the correct solution because the cycle-consistency loss always strives for a more expressive overall system.
	(This process is displayed in Fig.~\ref{fig:style-transfer-training-progress} on the right edge from top to bottom.)
	
	In the initial training cycle, the rendered image is constant because there are no known 3D objects and just the initially guessed sphere is rendered.
	When the reconstructions improve, the data generating process for the smooth renderings also improves, improving the style-transfer in return.
	In each training cycle, we train the domain adaptation for $1\,500$ iterations and then the reconstruction network for $2\,000$ iterations.
	We perform a total of $20$ cycles.
	
	The input, including background, can be reconstructed by the autoencoder because it can be compressed into high-frequency details as explained in detail by Chu~\etal~\cite{Chu2017CycleGANMasterSteganography}.
	
	In Supp.~Mat.~\ref{sub:style-variance}, we investigate the impact of variations in image brightness to our method.

	\subsection{Reconstructions from ShapeNet2Lego}
	
	In Fig.~\ref{fig:lego-recon}, we show single-view 3D reconstructions from photos of LEGO objects from a model trained on images with Minecraft backgrounds with our method.
	Here, even throughout different lighting conditions and backgrounds, 3D models can be reconstructed.
	We found that for the evaluated shapes, we achieve an average 3D IoU compared to the ground truth CAD models of $33.16\%$ which is lower than the performance of the original ShapeNet test set of $38.16\%$, but still performs in a comparable range given the domain gap compared to the training data.
	Comparing the reconstructed models in detail, we find that conditions like lightning or azimuth have an impact as can be seen in the reconstruction from image two, three, and four. Here, especially, the legs appear thinner under one projection (image two) than under another angle (image three).

	\section{Conclusion}
	\label{sec:discon}
	We proposed an adversarial style-transfer autoencoder pipeline to generalize state-of-the-art unsupervised single-view 3D geometry reconstruction methods. %
	To this end, we employed image-to-image translation-based adversarial domain adaptation to translate from the input image domain to the rendered image domain, which relieves the need for a photo-realistic renderer, since it does not require the assumption that the renderer can generate the input images.
	Experiments demonstrate that our style-agnostic approach can improve state-of-the-art methods in settings with uniform backgrounds and textured backgrounds.%
	\footnote{\textbf{Acknowledgements:} This work was supported by the IBM-MIT Watson AI Lab, the DFG in the SFB Transregio 161 ``Quantitative Methods for Visual Computing'' (Project-ID 251654672), and the Cluster of Excellence `Centre for the Advanced Study of Collective Behaviour'.}

	\clearpage%
	\printbibliography

	\clearpage
	\appendix

	\section{Experiment: Variance of Brightness}
	\label{sub:style-variance}
	
	As an additional experiment, we apply an artificial variation of brightness to the input and training images.
	By using extreme values, i.e., values that would generally even lie outside the valid range for images, we explore limitations of our approach.

	\paragraph{Type of Variation.}
	For brightness variation, we change the background color as well as the object brightness by random draws from a Gaussian distribution.
	Given standard deviation $\sigma$ and image $\mathrm{Img}$ as well as silhouette $\mathrm{Sil}$, we draw the variated image $\mathrm{Img}^\prime$ as follows:
	\begin{align}
		\mathrm{Img}^\prime 
		\sim 
		\mathrm{Img} \cdot (1 + \mathrm{Sil} \cdot \mathcal{N}(0,\sigma^2)) + |(1-\mathrm{Sil}) \cdot \mathcal{N}(0,\sigma^2)|
	\end{align}
	Note that, especially for larger $\sigma$, in most images the values actually exceed the range of values in the image ($[0, 1]$).
	We do \textit{not} clip the values to $[0, 1]$.
	
	\paragraph{Hyperparameters.}
	In this experiment, we use $400$ iterations for each style-transfer training to avoid instability due to the extreme variations and values in the data.
	
\begin{figure}[h]
		\centering
		\resizebox{\linewidth}{!}{
			\input{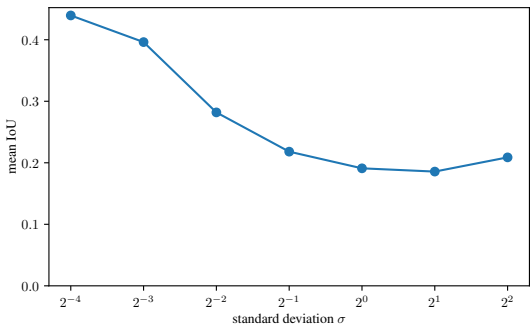}
		}
		\caption{
			Mean IoU on the ShapeNet data set for varying degrees of variance applied to brightness.
		}
		\label{fig:variance-results}
	\end{figure}

	As shown in the plot in Fig.~\ref{fig:variance-results}, for increasing variance of brightness, the accuracy drops, however, even for an extreme standard deviation of $4$, the model still learns and does not break down entirely.

	\section{Implementation Details}

	\subsection{Image-to-Image Networks and Discriminator}
	\label{sec:implementation-image-to-image}
	Our basic building block performs convolution with filter size $4\times 4$ and stride~$2$, followed by batch normalization and Leaky-ReLU with a slope parameter of~$\alpha=0.2$.
	We abbreviate this standard layer by~\layers{Ck}, where $k$ is the number of filters.
	The networks $\itr$ and $\rti$ are both symmetric residual networks, where the first
	half is defined by the chain of blocks
	\layers{C64-C128-C256-C512-C512}.
	As final activation, we add a $\tanh$.
	The discriminator $\discr$ consists of the residual blocks
	\layers{C64-C128-C256-C1} followed by a logistic sigmoid activation.

	\subsection{3D-Geometry Reconstruction Network}
	\label{sub:reconstructor}\label{sec:pix2vex}
	
	The architecture is based on convolutional layers (\layers{Ck}) with \layers{k} filters of size $5\times5$, a stride of 2, a padding of 2, and batch normalization.
	Further, it uses fully connected layers (\layers{Fk}), where~\layers{k} denotes the output dimension.
	The activation function for all layers except the last one is ReLU.
	With these definitions, the architecture can be written as \layers{C64-C128-C256-F1024-F1024-F512-F1024-} \layers{-F2048-F1929}.

	\subsection{Ratio between Style and Content Loss}
	
	\setlength{\tabcolsep}{4pt}
	\begin{table}[h]
		\caption{
			Comparing ratios between the cycle-consistency content loss and the adversarial style loss.
		}
		\centering
		\small
		\begin{tabular}{rrc} 
			\hline \noalign{\smallskip}
			Style loss & Cycle-consistency loss & IoU~Acc.  \\ 
			\hline \noalign{\smallskip}
			1 & 100 & 0.2525  \\ 
			1 & 200 & 0.3458  \\ 
			1 & \textbf{400} & 0.3816  \\ 
			1 & 800 & 0.2925 \\
			1 & 1\,600 & 0.2844 \\ 
			\hline
		\end{tabular}
	\end{table}
	\setlength{\tabcolsep}{1.4pt}

	\section{3D Reconstruction Examples}
\begin{figure*}
		\centering
		\includegraphics[width=.85\textwidth]{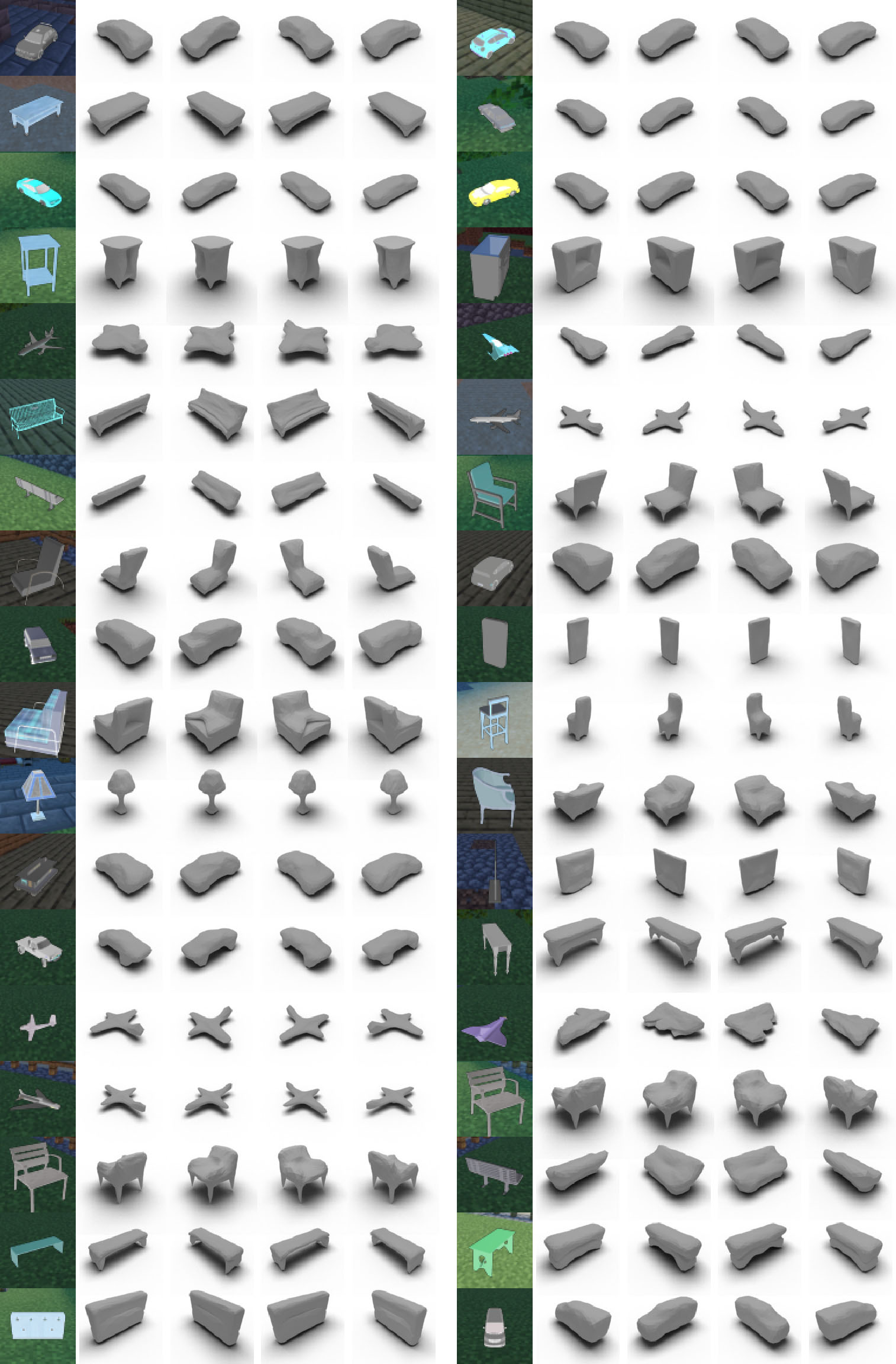}
		\caption{3D Reconstruction Examples 1\label{fig:recon-ex1}}
	\end{figure*}

\begin{figure*}
		\centering
		\includegraphics[width=.85\textwidth]{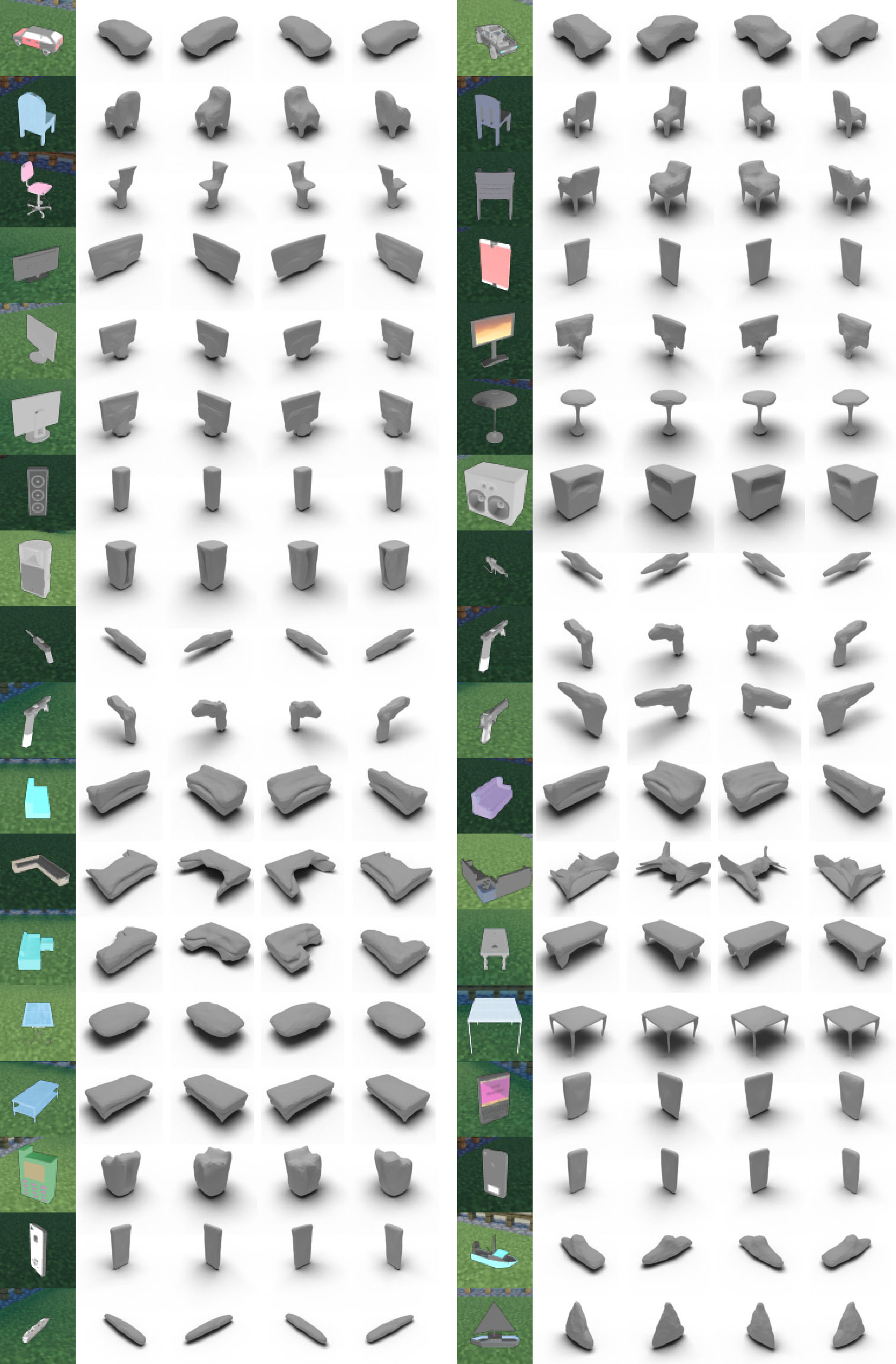}
		\caption{3D Reconstruction Examples 2\label{fig:recon-ex2}}
	\end{figure*}

	In Fig.~\ref{fig:recon-ex1}--\ref{fig:recon-ex2}, we present further 3D mesh reconstruction examples.

	\section{Training of the Style-Transfer Examples}
	\label{sm:style-transfer-examples}
	
	As in Fig.~6 of our work, we will show here the training process for 12 input images.
	Note that the first five rows display the first five training cycles while the \textit{sixth row displays the tenth training cycle}.
	All other parameters are as in Fig.~6 of our work.
	Note that this covers only half of the entire training as we used 20 training cycles in our experiments.
	
	\newpage
	
	\pagestyle{empty}
	
	\begin{figure*}
		\centering
		\includegraphics[width=.8\textwidth]{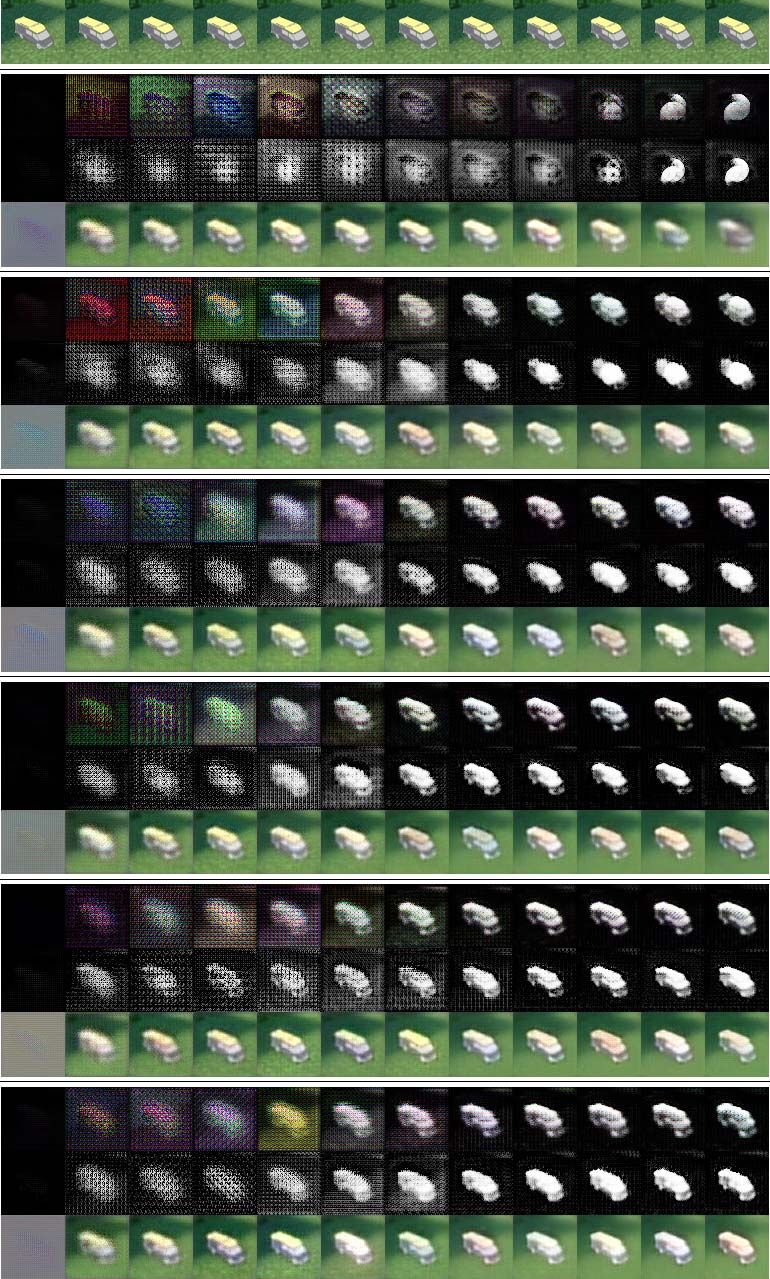}
		\caption{Training of the Style-Transfer Example 1}
	\end{figure*}
	
	\begin{figure*}
		\centering
		\includegraphics[width=.8\textwidth]{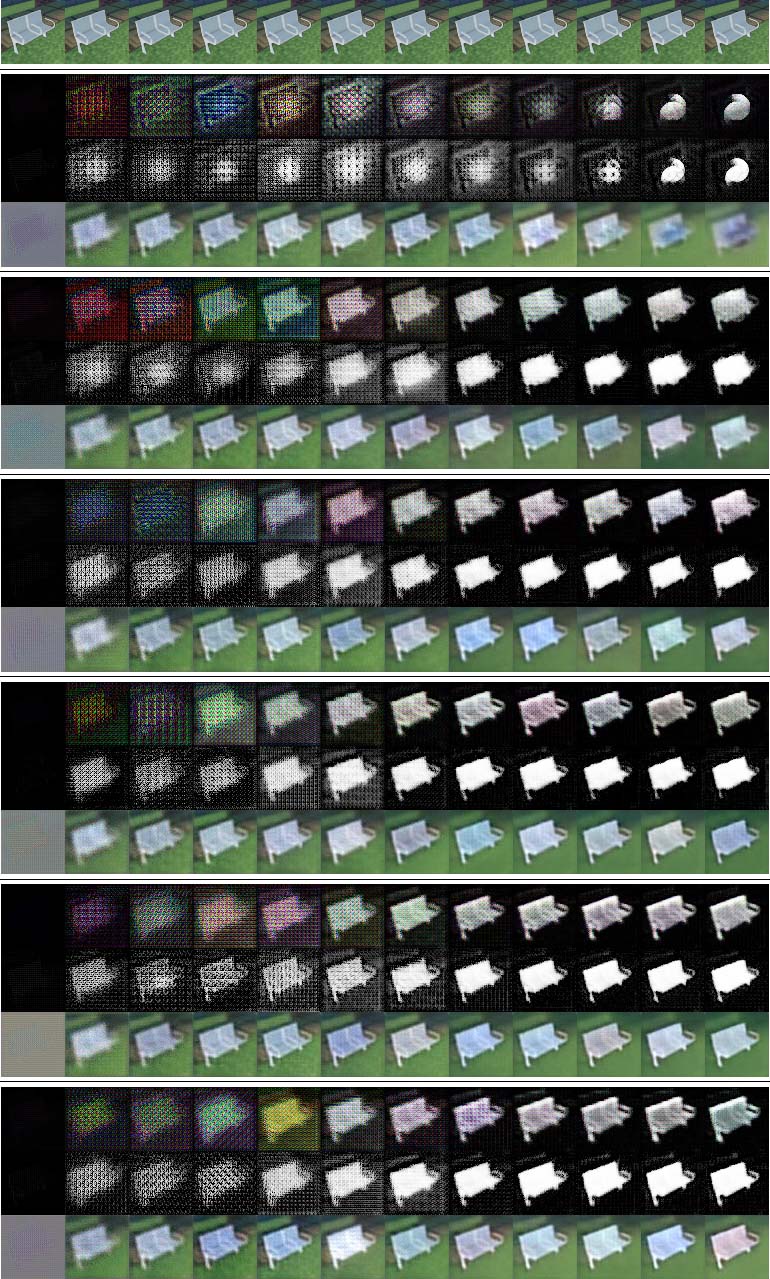}
		\caption{Training of the Style-Transfer Example 2}
	\end{figure*}
	
	\begin{figure*}
		\centering
		\includegraphics[width=.8\textwidth]{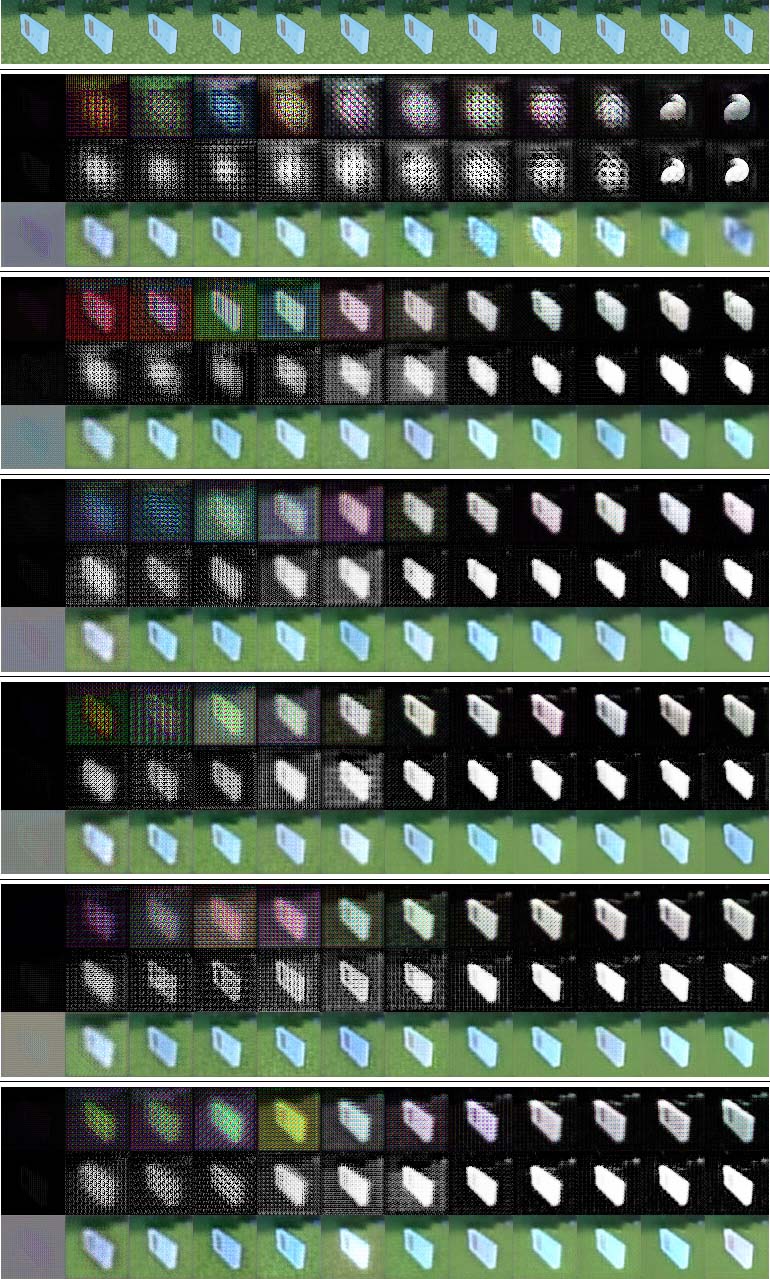}
		\caption{Training of the Style-Transfer Example 3}
	\end{figure*}
	
	\begin{figure*}
		\centering
		\includegraphics[width=.8\textwidth]{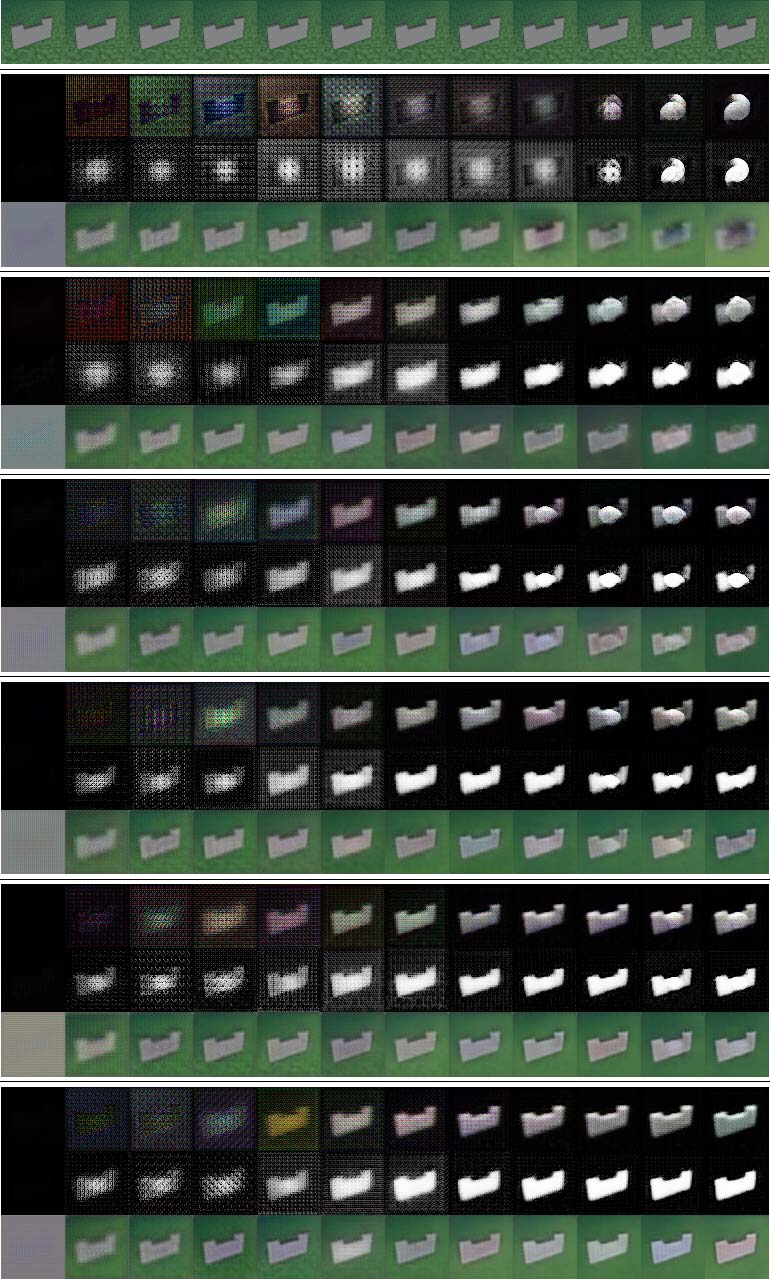}
		\caption{Training of the Style-Transfer Example 4}
	\end{figure*}
	
	\begin{figure*}
		\centering
		\includegraphics[width=.8\textwidth]{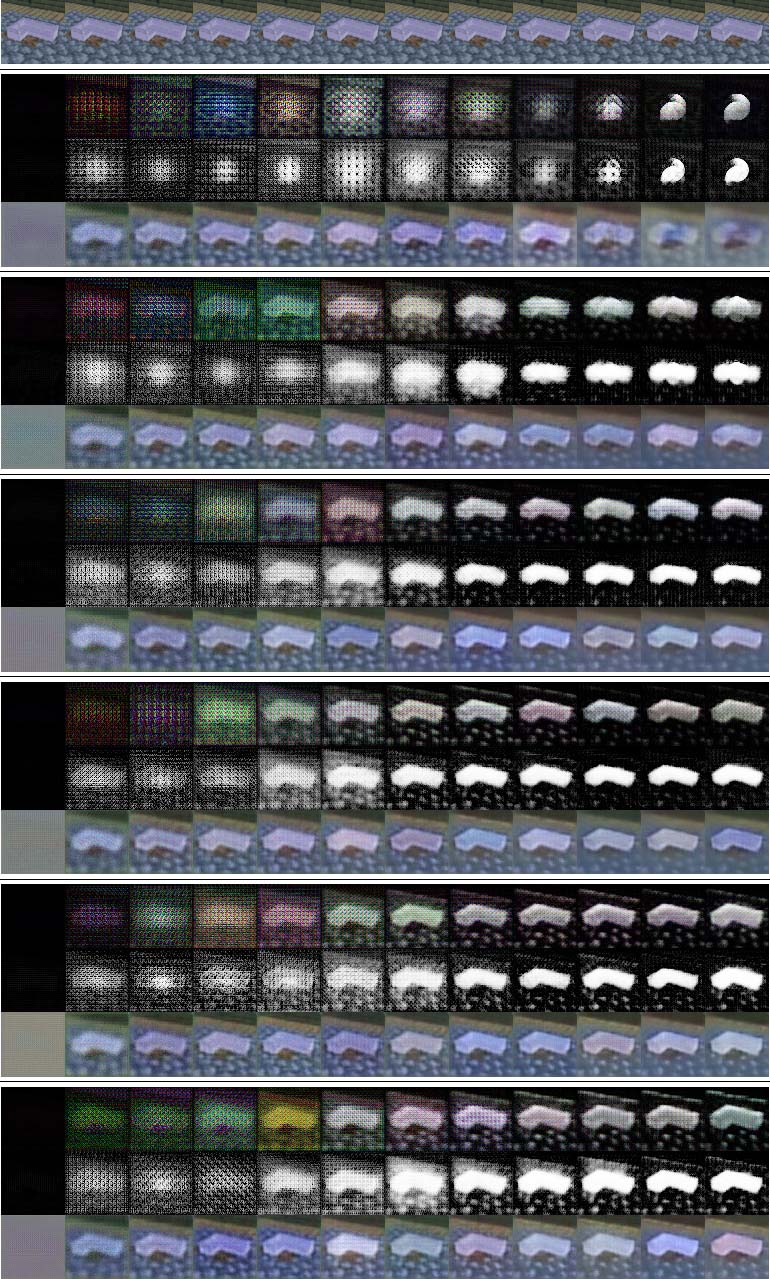}
		\caption{Training of the Style-Transfer Example 5}
	\end{figure*}
	
	\begin{figure*}
		\centering
		\includegraphics[width=.8\textwidth]{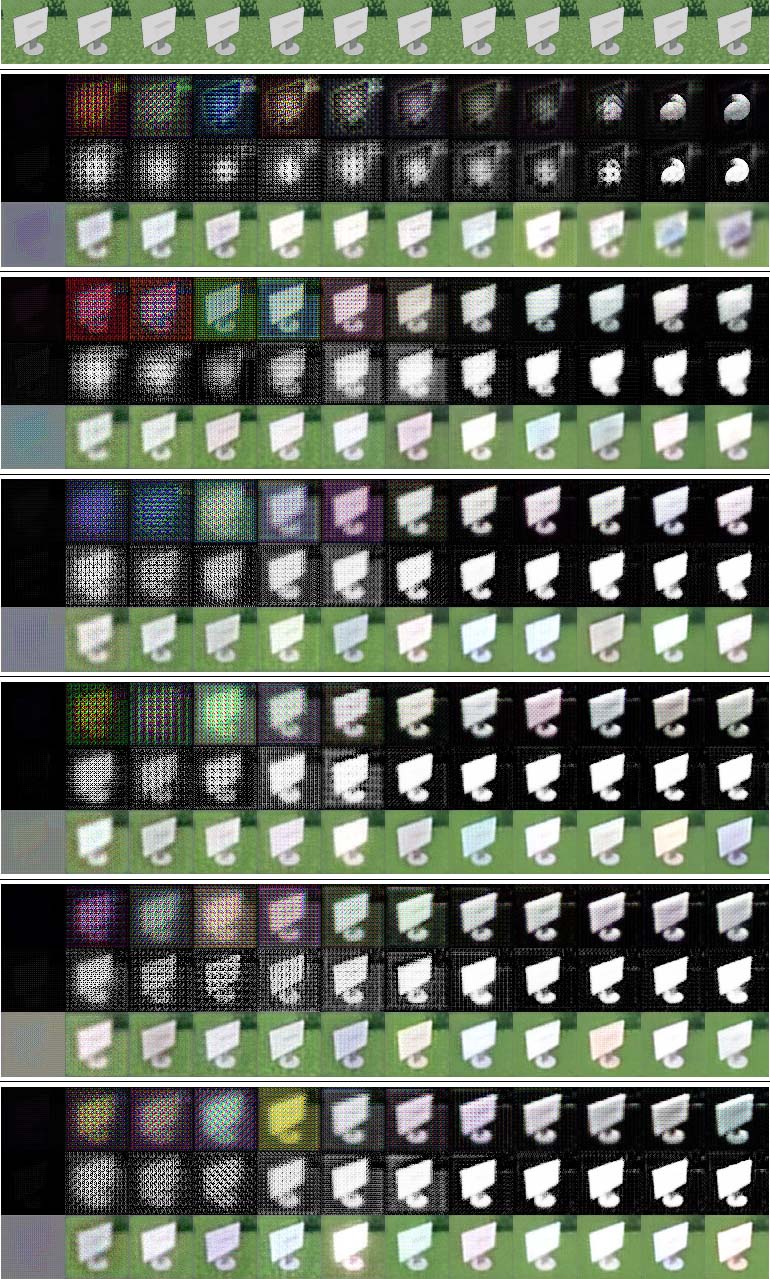}
		\caption{Training of the Style-Transfer Example 6}
	\end{figure*}
	
	\begin{figure*}
		\centering
		\includegraphics[width=.8\textwidth]{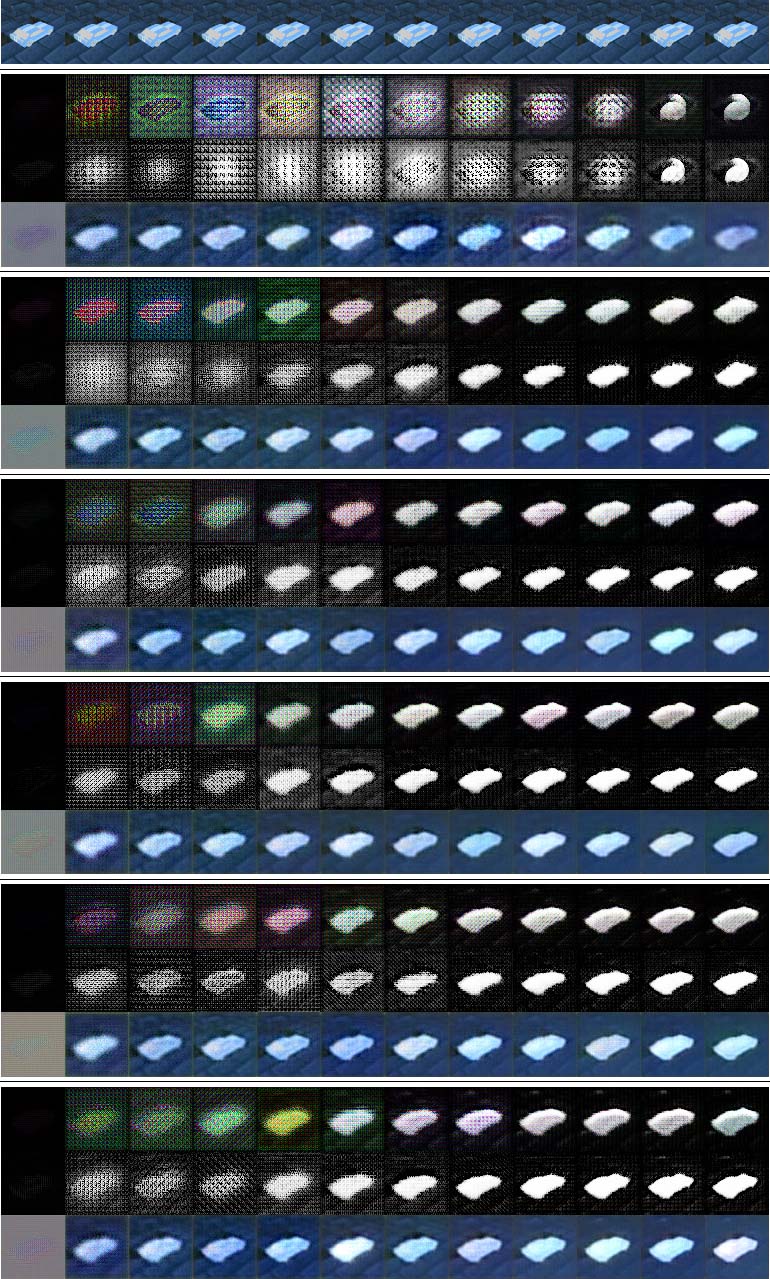}
		\caption{Training of the Style-Transfer Example 7}
	\end{figure*}
	
	\begin{figure*}
		\centering
		\includegraphics[width=.8\textwidth]{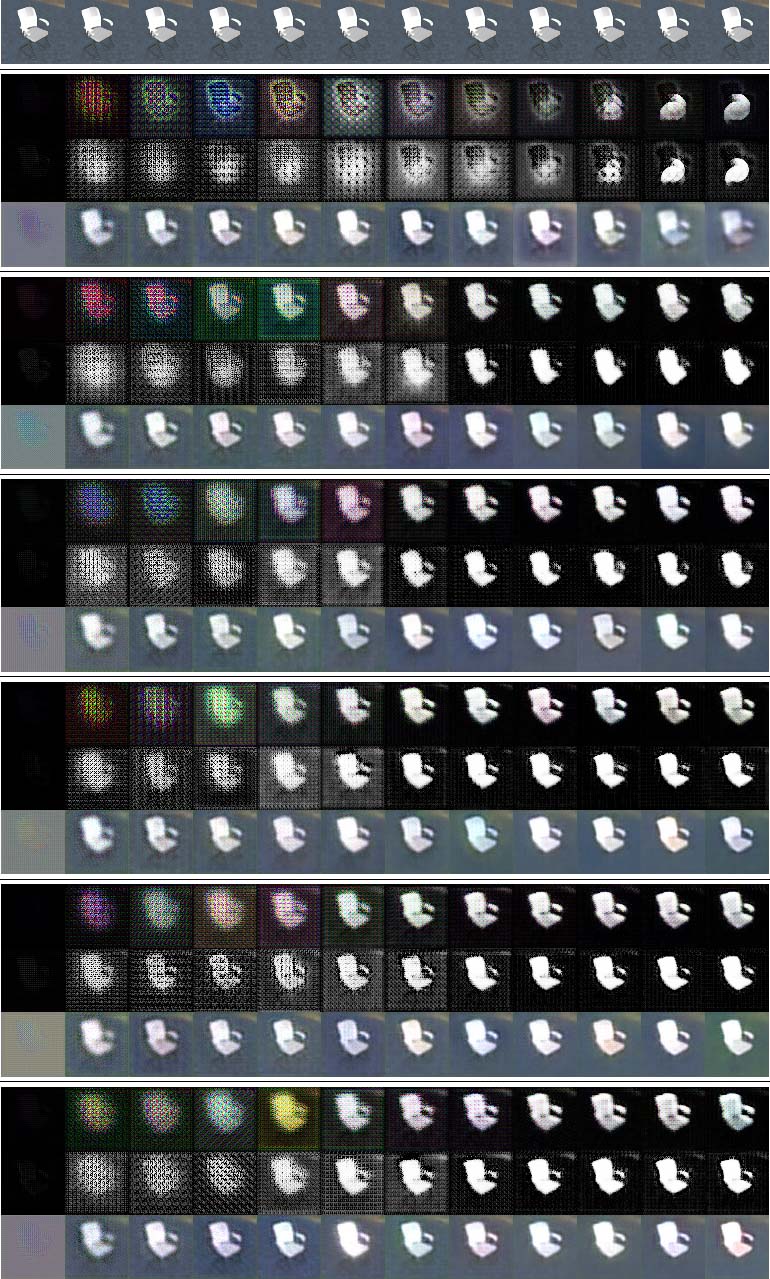}
		\caption{Training of the Style-Transfer Example 8}
	\end{figure*}
	
	\begin{figure*}
		\centering
		\includegraphics[width=.8\textwidth]{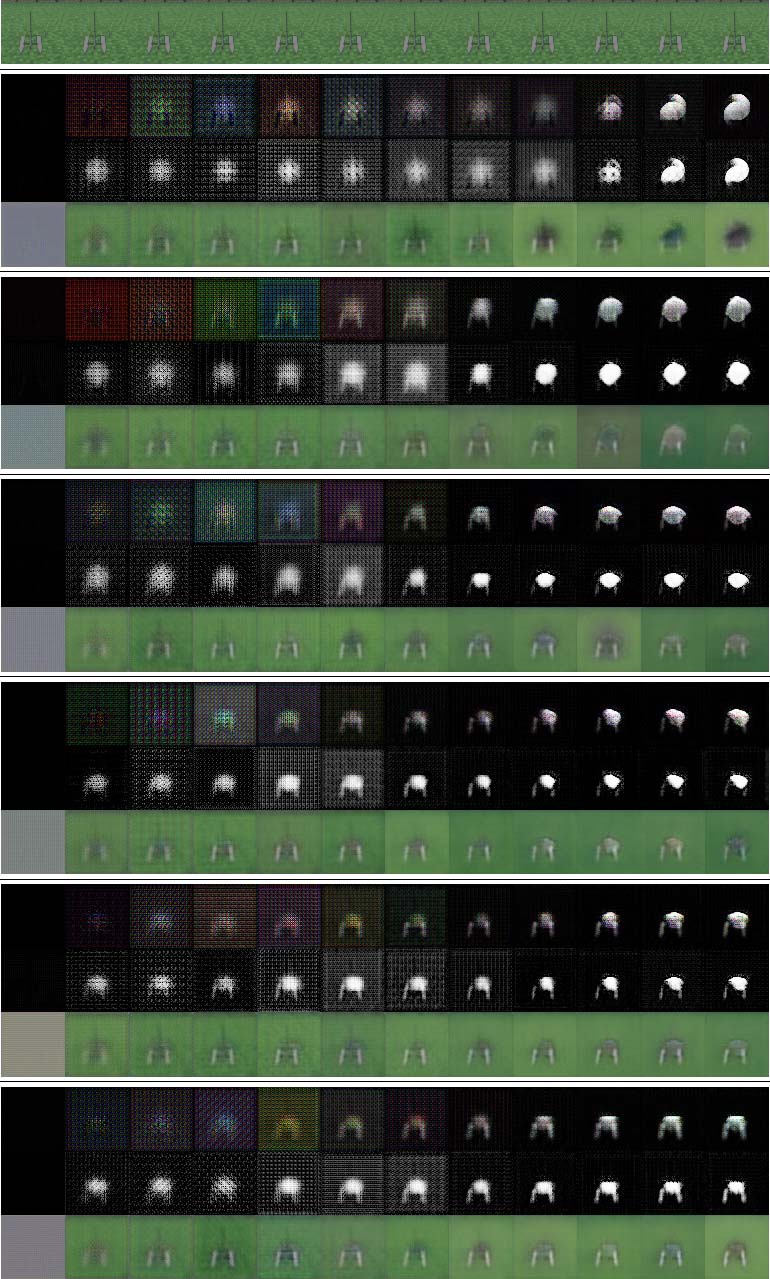}
		\caption{Training of the Style-Transfer Example 9}
	\end{figure*}
	
	\begin{figure*}
		\centering
		\includegraphics[width=.8\textwidth]{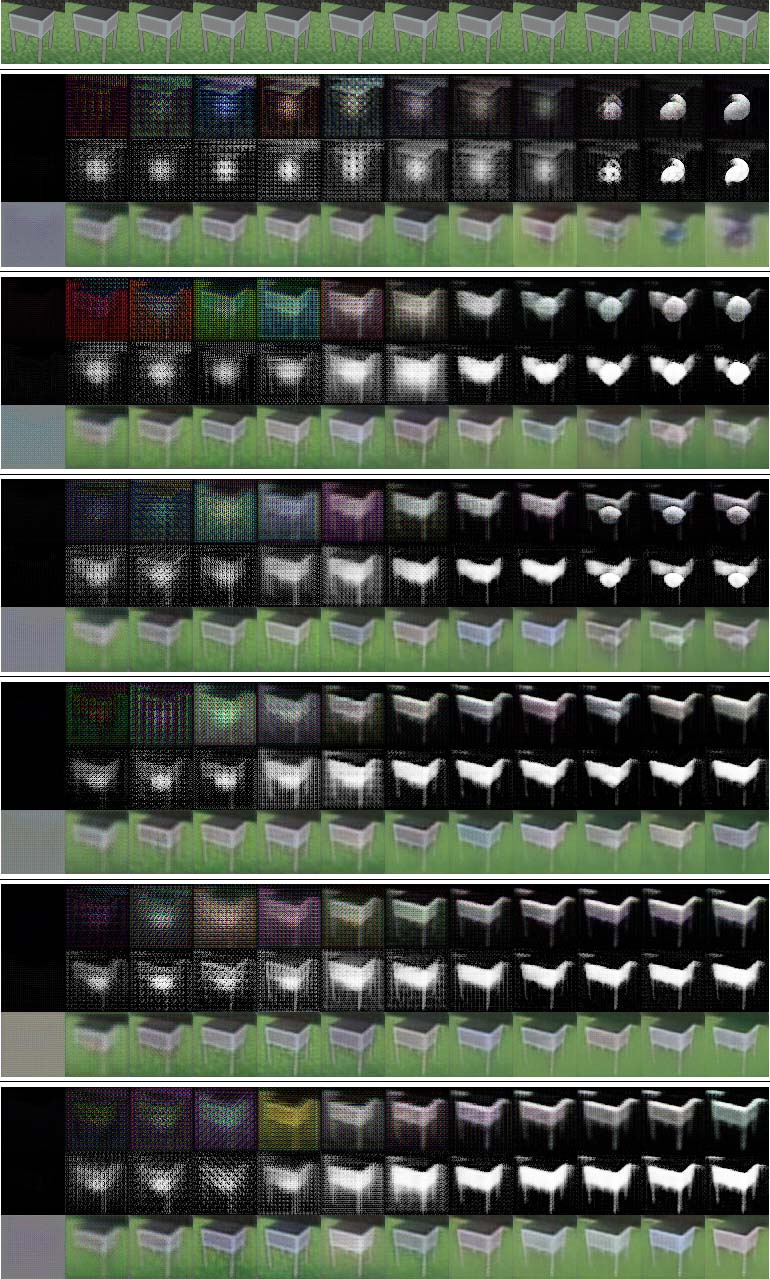}
		\caption{Training of the Style-Transfer Example 10}
	\end{figure*}
	
	\begin{figure*}
		\centering
		\includegraphics[width=.8\textwidth]{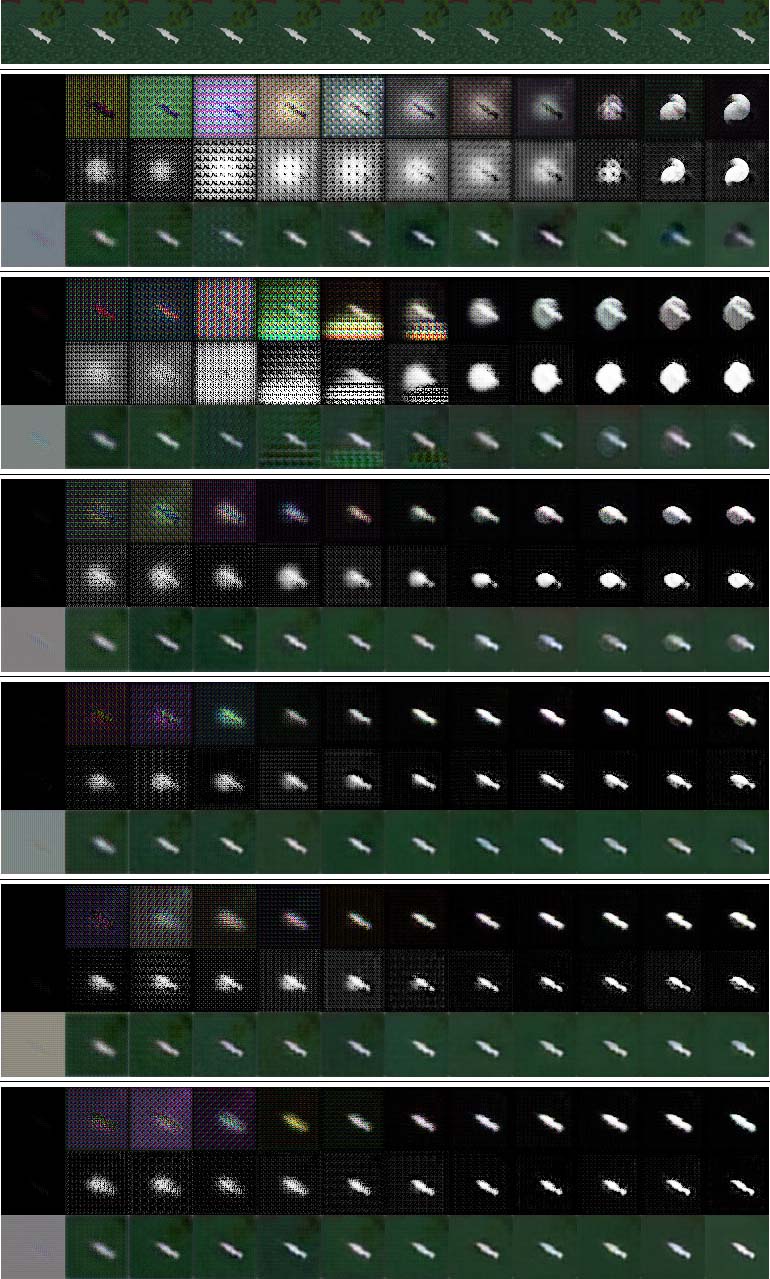}
		\caption{Training of the Style-Transfer Example 11}
	\end{figure*}
	
	\begin{figure*}
		\centering
		\includegraphics[width=.8\textwidth]{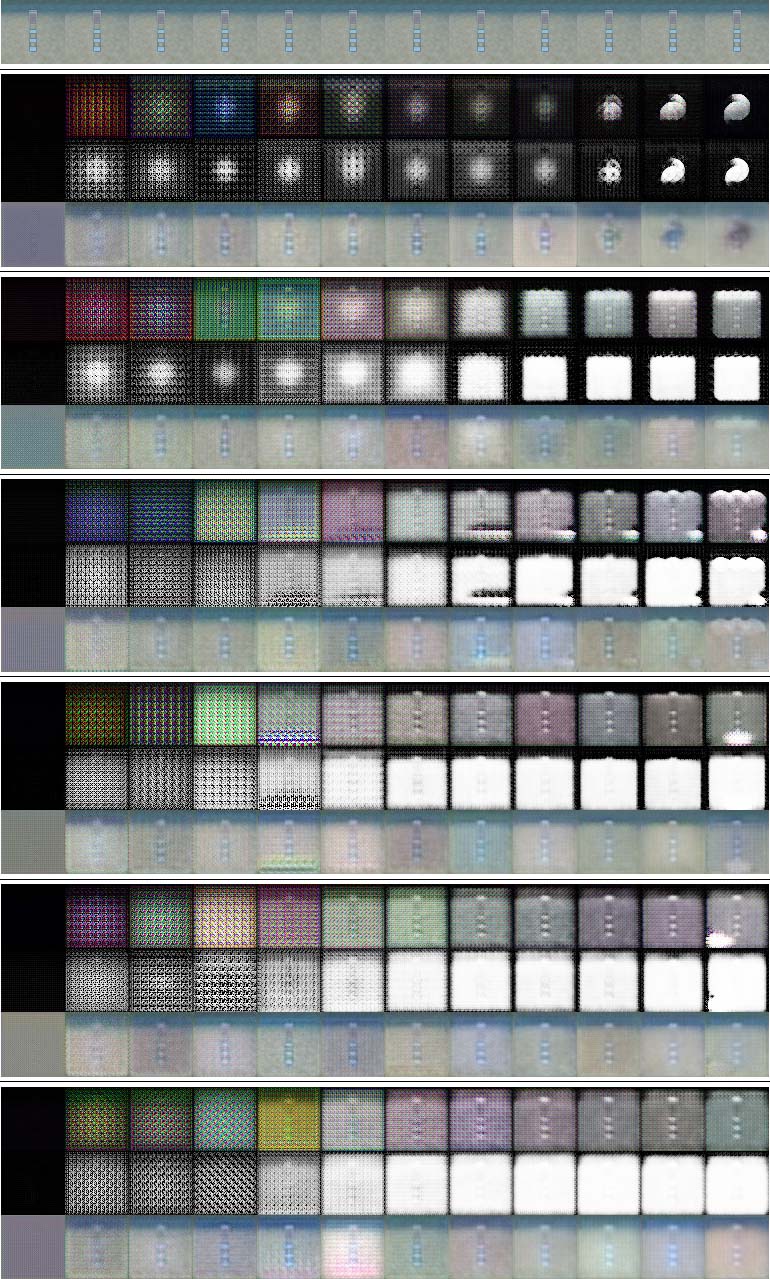}
		\caption{Training of the Style-Transfer Example 12}
	\end{figure*}

\end{document}